\documentclass[sigconf,natbib,screen=true]{acmart}

\acmSubmissionID{1033}

\usepackage{acronym}
\usepackage[skip=0pt]{caption}
\usepackage[inline]{enumitem}
\usepackage{booktabs}
\usepackage{makecell}
\usepackage{multirow}
\usepackage{subcaption}
\usepackage{xspace}
\usepackage{adjustbox}
\usepackage{bbding}
\usepackage{pifont}
\usepackage{caption}
\usepackage{graphicx}
\usepackage{float} 
\usepackage{subcaption}

\AtBeginDocument{%
  \providecommand\BibTeX{{%
    \normalfont B\kern-0.5em{\scshape i\kern-0.25em b}\kern-0.8em\TeX}}}

\newcommand{\yifei}[1]{[\textcolor{blue}{Yifei: #1}]}

\acrodef{DS}{dialogue system}
\acrodef{TDS}{task-oriented dialogue system}
\acrodef{ReDial}{recommendation dialogues}
\acrodef{IR}{information retrieval}
\acrodef{AMT}{Amazon Mechanical Turk}
\acrodef{HIT}{human intelligence task}
\acrodef{CQA}{community question answering}
\acrodef{MQC}{multimodal query clarification}
\acrodef{ROI}{region of interest}
\acrodef{QL}{query likelihood}
\acrodef{MRR}{mean reciprocal rank}
\acrodef{LTR}{learning to rank}

\newcommand{\OurData}{Melon\xspace}
\newcommand{\OurModel}{Marto\xspace}
\newcommand{\OurModelNS}{Marto}
\newcommand{\TOR}{TOR\xspace}
\newcommand{\MUR}{MUR\xspace}
\newcommand{\POS}{VEQ\xspace}
\newcommand{\NEG}{TEQ\xspace}

\newcommand{\header}[1]{\vspace*{1mm}\noindent\textbf{#1}.}

\newcommand{\cmark}{\ding{51}}%
\newcommand{\xmark}{\ding{55}}%

\copyrightyear{2024}
\acmYear{2024}
\setcopyright{acmlicensed}\acmConference[WWW '24]{Proceedings of the ACM Web Conference 2024}{May 13--17, 2024}{Singapore, Singapore}
\acmBooktitle{Proceedings of the ACM Web Conference 2024 (WWW '24), May 13--17, 2024, Singapore, Singapore}
\acmDOI{10.1145/3589334.3645483}
\acmISBN{979-8-4007-0171-9/24/05}
\author{Yifei Yuan}
\affiliation{%
  \institution{University of Copenhagen}
  \city{Copenhagen}
  \country{Denmark}}
\email{yiya@di.ku.dk}

\author{Clemencia Siro}
\affiliation{%
  \institution{University of Amsterdam}
  \city{Amsterdam}
  \country{The Netherlands}}
\email{c.n.siro@uva.nl}

\author{Mohammad Aliannejadi}
\affiliation{%
  \institution{University of Amsterdam}
  \city{Amsterdam}
\country{The Netherlands}}
\email{m.aliannejadi@uva.nl}

\author{Maarten de Rijke}
\affiliation{%
  \institution{
  University of Amsterdam}
  \city{Amsterdam}
  \country{The Netherlands}}%
\email{m.derijke@uva.nl}

\author{Wai Lam}
\affiliation{
\institution{The Chinese University of Hong Kong}
\city{Hong Kong}
\country{Hong Kong, SAR}}
\email{wlam@se.cuhk.edu.hk}

\allowdisplaybreaks

\begin{document}

\title[Asking Multimodal Clarifying Questions in Mixed-Initiative Conversational Search]{Asking Multimodal Clarifying Questions\\ in Mixed-Initiative Conversational Search}

\begin{abstract}
In mixed-initiative conversational search systems, clarifying questions are used to help users who struggle to express their intentions in a single query.
These questions aim to uncover user's information needs and resolve query ambiguities.
We hypothesize that in scenarios where multimodal information is pertinent, the clarification process can be improved by using non-textual information. Therefore, we propose to add images to clarifying questions and formulate the novel task of asking multimodal clarifying questions in open-domain, mixed-initiative conversational search systems. 
To facilitate research into this task, we collect a dataset named \OurData that contains over 4k multimodal clarifying questions, enriched with over 14k images. 
We also propose a multimodal query clarification model named \OurModel and adopt a prompt-based, generative fine-tuning strategy to perform the training of different stages with different prompts. Several analyses are conducted to understand the importance of multimodal contents during the query clarification phase. Experimental results indicate that the addition of images leads to significant improvements of up to 90\% in retrieval performance when selecting the  relevant images. Extensive analyses are also performed to show the superiority of \OurModel compared with discriminative baselines in terms of effectiveness and efficiency.  
\end{abstract}

\ccsdesc[100]{Information systems~Evaluation of retrieval results}

\keywords{
Query clarification, 
Multimodal query understanding}

\maketitle

\acresetall

\section{Introduction}


Traditional search systems often struggle to provide relevant results that meet a user's information need when they encounter incomplete or ambiguous queries. 
Query clarification has emerged as a promising approach to address this challenge~\cite{Zamani2020GeneratingCQ,AliannejadiSigir19}.
It enables systems to interact with users to clarify their information needs before presenting search results~\cite{Hancock2019LearningFD,Rao2018LearningTA}. 
While previous research has focused primarily on unimodal interactions, the growing popularity of smart displays underscores the increasing demand for multimodal communication~\cite{Liao2021MMConvAE}.
Specifically, the use of multimodal information improves effectiveness in various \ac{IR} tasks, including image-based search~\cite{Xie2018ConstructingAI}, e-commerce cross-modal retrieval~\cite{Gao2020FashionBERTTA}, and fashion recommendation~\cite{Yin2019EnhancingFR}. In this context, multimodal information not only provides a visually appealing experience to the users but also enhances both the system and user performance, allowing for the integration of complementary information across multiple modalities. 

\begin{figure}[!h]
    \centering
    \vspace*{-.5mm}
\includegraphics[width=0.95\linewidth]{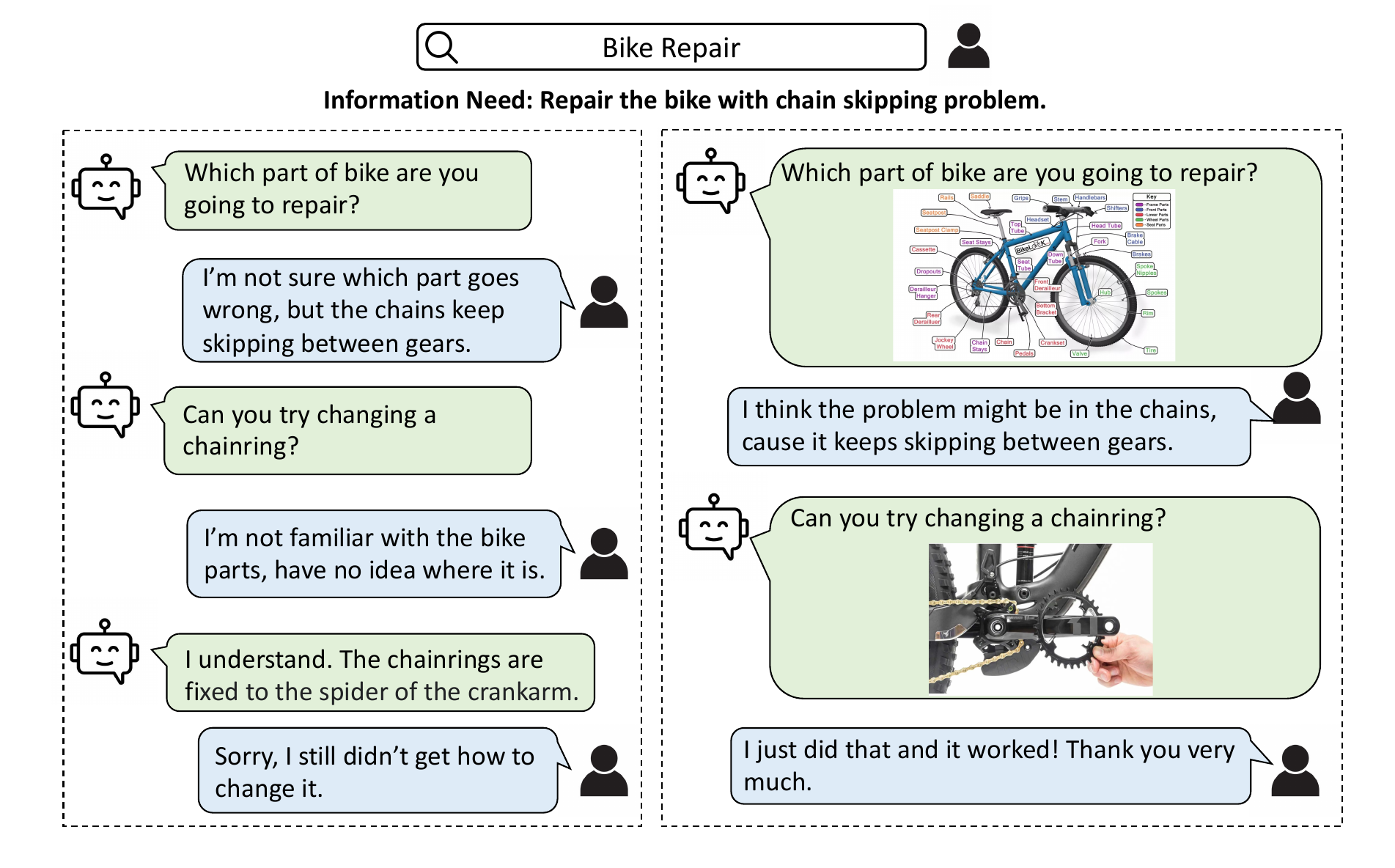}
    \caption{An example of incorporating multimodal information into the query  clarification phase. }
    \label{fig:exmaple}
    \vspace*{-.5mm}
\end{figure}

\header{Motivation}
As the interest in multimodal information-seeking conversations continues to grow~\cite{Deldjoo2021TowardsMC,tsagkias-2020-challenges}, we study the use of multimodal information in query clarification. 
We hypothesize that the addition of relevant images to clarifying questions within a conversational context can enhance the user experience and performance, providing a rich source of information that augments textual input. Consequently, this should lead to increased awareness and understanding of the information need and various aspects of it such as domain knowledge. Fig.~\ref{fig:exmaple} shows two example conversations --- without and with images.
We see that when a user inputs the query ``bike repair'' with the underlying information need of ``repair a bike with a chain skipping problem,'' presenting a clarifying question accompanied by an image of different parts of a bike can provide a visual representation of the intended query, allowing the user to see the object they are searching for. 
The potential benefits of this approach are manifold, including: 
\begin{enumerate*}[label=(\roman*),nosep]
    \item displaying the appropriate set of images with clarifying questions reassures the user that the system has correctly understood their request;
    \item showing images can help the user acquire a more detailed and complete comprehension of the search topic domain;
    and
    \item it could enable the user to engage in discussions and inquiries, pertaining to visual content, such as a specific part of a bike. 
\end{enumerate*}
Ultimately, these benefits help ensure that the retrieved results are better aligned with the user's actual information needs, leading to improved retrieval effectiveness and efficiency.  
Besides, as visual content provides a more engaging and immersive interaction, it leads to increased user engagement~\cite{Liao2018KnowledgeawareMD} and satisfaction with the search experience~\cite{Deldjoo2021TowardsMC}. 
E.g., users may have difficulty understanding the conversation when the system relies on text descriptions to explain the process of a bike repair.

\header{Research goals}
In this study, we formulate a novel task of asking multimodal clarifying questions in open-domain mixed-initiative conversational search systems. We define \ac{MQC} as \textit{asking clarifying questions enriched with images in a text-dominant conversation} and present the workflow depicted in Fig.~\ref{fig:workflow}.
Initially, when a user submits a search query with some intention that is concealed from the system, the system retrieves relevant documents and evaluates their appropriateness for presentation to the user. 
For ambiguous queries where the system struggles to understand the user's intent, a clarifying question is asked.
Our approach employs a classifier to determine whether to include images in the clarifying question. 
Upon confirmation (this type of question is defined as a \textbf{multimodal clarifying question}), images are retrieved and displayed to the user with the clarifying question. 
In this case, visual information serves as a valuable resource, enhancing user comprehension and supporting informed responses. 
Therefore, combined with prior context, it can deliver precise and relevant retrieval results.

To advance research into the proposed task, following the workflow described above, we collect a new dataset \emph{\OurData} (\textbf{m}ulti\-modal qu\textbf{e}ry c\textbf{l}arification conversati\textbf{on}s), based on the text-only clarifying question dataset ClariQ~\citep{Aliannejadi2020ConvAI3GC}. 
\OurData contains over 4k multimodal clarifying questions, which are enriched with over 14k images. With \OurData, we investigate the following research questions: \textbf{RQ1}: \textit{What is the effect of including multimodal content in the document retrieval task?} To address \textbf{RQ1}, inspired by generative retrieval techniques in the text-only retrieval domain~\cite{Lee2022ContextualizedGR,DeCao2020AutoregressiveER}, we propose \OurModel (\textbf{m}ultimod\textbf{a}l que\textbf{r}y clarifica\textbf{t}ion m\textbf{o}del) and compare it with several state-of-the-art unimodal models. \OurModel is based on a multimodal generative framework and adopts a multi-task fine-tuning approach to train different stages of \OurModel with different prompts. 
We initially train \OurModel to generate `true' or `false' labels to assess if a given clarifying question is multimodal. For `true' cases, it selects an image based on its similarity to the question. We also develop a generative document retrieval method where all questions with or without images are trained to generate document IDs. By leveraging pre-trained knowledge, this approach equips \OurModel with enhanced retrieval ability in an end-to-end manner. Our analyses indicate that adding images can lead to up to 90\% improvements in performance.

In \textbf{RQ2}, we address: \textit{How does the attachment of different images affect the performance of multimodal query clarification in terms of retrieval accuracy?} We answer this question by conducting a performance comparison between \OurModel and its variants by attaching different images and several multimodal baselines. We show that selecting the relevant image is beneficial to the retrieval performance compared with attaching a random one. Our findings provide important insights into the design of \OurModel and aid the development of more advanced conversational search systems that take advantage of the rich information in multimodal interactions.

To further assess the performance of \OurModel and address \textbf{RQ3}: \textit{Are generative clarification models more effective and efficient for document retrieval?}, we conduct experiments and analyze the training progress in terms of loss and validation of competitive generative and discriminative clarification models. Our results reveal that Marto, in comparison to its discriminative counterpart VisualBERT, exhibits considerably reduced training time (0.77 vs. 8.26) and inference time (0.67 vs. 1.13). Thus indicating the superiority of generative clarification models in document retrieval .

Finally, to understand the significance of multimodal content in the query clarification process and its potential impact on users, we address \textbf{RQ4}:
\textit{How does the inclusion of multimodal clarifying questions impact the user response?} To answer this question, we perform analyses from both dataset and model perspectives. Our findings show that using images during the clarification phase leads to more contextualized answers. This results in richer semantic information compared to datasets that only feature text-based clarifications.





\header{Contributions}
Our main contributions are as follows:
\begin{itemize}[leftmargin=*,nosep]
    \item We define a novel task of \ac{MQC} in a mixed-initiative conversational search system, which adds image information in clarifying questions to improve the downstream document retrieval task. 
    \item To facilitate the offline evaluation of the task, we benchmark a large-scale dataset called \OurData and propose \OurModel for representation learning of multimodal clarifying questions.\footnote{We release the dataset and code in \url{https://github.com/yfyuan01/MQC/}. To preserve the copyright of the images we will only share the public URLs to the images. Also, we will ask the researchers to sign a license agreement.}
    \item Extensive analyses are performed to explore the impact of multimodal clarifying questions on user response and their role in improving retrieval performance.
\end{itemize}

\vspace*{-2mm}
\section{Related Work}


\header{Query clarification in mixed-initiative search systems}
Query clarification refers to the process of improving a search query by adding more context or details to it. In recent years, query clarification has become an essential task in various domains, such as entity disambiguation~\cite{Coden2015DidYM}, voice~\cite{Kiesel2018TowardVQ}, dialogue~\cite{Boni2003AnAO,Quintano2008QuestionAnsweringCD}, question answering~\cite{Braslavski2017WhatDY,Xu2019AskingCQ,Rao2018LearningTA}, recommendation~\cite{Zhang2018TowardsCS,Christakopoulou2016TowardsCR}. 
In mixed-initiative search systems where the initiative shifts back and forth between agents and users~\cite{Hearst1999TrendsC,Allen1999MixedinitiativeI}, asking clarifying questions has received considerable attention~\cite{Zamani2020AnalyzingAL,Sekulic2021UserEP,Hashemi2020GuidedTL,Owoicho2023ExploitingSU}. Efforts have been made to investigate the role of clarifying questions in mixed-initiative systems, recognizing their potential to improve search quality and user experience~\cite{Vakulenko2020AnAO,Meng2021InitiativeAwareSL,Aliannejadi2021AnalysingMI}. 
To explore when to ask clarification questions, the TREC CAsT 2022 track included a task where the system can either take the initiative by posing questions or generating a response.\footnote{\url{https://www.treccast.ai/}}
Resources have been proposed to facilitate the offline evaluation of such systems. For example, MIMICS is a large-scale dataset sampled from the Bing query logs~\cite{Zamani2020MIMICSAL,Tavakoli2022MIMICSDuoO}.  
In open-domain information-seeking conversations, Qulac is a clarification dataset that contains clarifying questions and answers~\cite{AliannejadiSigir19}. 
ClariQ expands Qulac to a multi-turn format and has a larger scale~\cite{Aliannejadi2020ConvAI3GC}. 
Despite the progress in mixed-initiative search clarification, there is a significant gap in our understanding of how to incorporate multimodal information in query clarification. 

\header{Multimodal IR}
Multimodal IR involves integrating multimodal query processing techniques to effectively capture users' diverse search intent~\cite{Mouro2015MultimodalMI,Srihari2000AMF,Bokhari2013MultimodalIR,Narayan2003StagingTF}, whose techniques can be applied in various IR scenarios~\cite{Saraiva2016AMQ,Tautkute2021IWT,Datta2017MultimodalRU,Gao2020FashionBERTTA,Xie2018ConstructingAI,Yin2019EnhancingFR,Yuan2021ConversationalFI}. 
In mixed-initiative systems, multimodal content is leveraged to enhance retrieval efficiency~\cite{Gao2020FashionBERTTA,Zou2022UsersMC,Deldjoo2021TowardsMC} and improve user experience~\cite{MurrugarraLlerena2018ImageRW,Ma2021MixedModalityII}. Inspired by the success of generative LLMs~\cite{Raffel2019ExploringTL,Brown2020LanguageMA}, a set of multimodal pre-trained generative models have been proposed recently~\cite{Zhang2021VinVLRV,Li2020OscarOA,Zhou2019UnifiedVP,DBLP:conf/icml/ChoLTB21}. 
Subsequent efforts have proven their effectiveness in IR tasks such as query reformulation~\cite{Yuan2022McQueenAB}, question answering~\cite{Chang2021WebQAMA,Talmor2021MultiModalQACQ}, cross-modal retrieval~\cite{Radford2021LearningTV}. 
Among them, VLT5~\cite{DBLP:conf/icml/ChoLTB21} is the state-of-the-art  multimodal pretrained model and shows strong performance, especially in the image caption task. 
We incorporate images into the clarification phase of mixed-initiative conversations and develop our model using VLT5 as the base model.

\header{Generative IR}
Generative retrieval, unlike traditional IR systems that follow an ``index-retrieve-then-rank'' pipeline,  generates identifiers of relevant documents in an end-to-end way~\cite{Chen2022CorpusBrainPA,Chen2023AUG,Chen2023UnderstandingDS}. 
Subsequent work has explored the generation of document titles for entity linking~\cite{DeCao2020AutoregressiveER}, fact-checking~\cite{Chen2022GEREGE}, and recommendation tasks~\cite{Wang2023GenerativeRT}.
Efforts have also been made to investigate and compare the effect of different document identifiers~\cite{Tay2022TransformerMA,Chen2023AUG,Lee2022ContextualizedGR}. 
E.g., \citet{Bevilacqua2022AutoregressiveSE} leverage n-grams as possible identifiers. 
\citet{Wang2022ANC} propose generating the document id and improve the system with a prefix-aware decoder.  
Motivated by the success of this context~\cite{DBLP:conf/icml/ChoLTB21,Radford2021LearningTV,OpenAI2023GPT4TR}, we propose \OurModel, the first multimodal generative retrieval model that takes multimodal information as input and generates the corresponding document identifier.

\vspace*{-2mm}
\section{Task Definition and Workflow}
\label{sec:multimodal-clarificaiton}
In a search system, the user submits a query with a latent intention or information need. The system's goal is to accurately predict this intention and retrieve relevant results. 
To achieve this, the system can use multimodal clarification questions, which are meant to help capture the user's needs.
We define the \acf{MQC} task as follows: \textit{multimodal query clarification refers to scenarios where the user-system interactions are mainly in text, while the clarifying questions are potentially enriched with multiple images.}
The main assumption is that the images
\begin{enumerate*}[label=(\roman*)]
\item are related to the conversation topic and the posed clarifying question, 
and 
\item provide further contextual information on the topic. 
\end{enumerate*}
E.g., in Fig.~\ref{fig:exmaple}, if a user is not familiar with different parts of a bike, attaching an image to the clarifying question can aid in their comprehension and provide additional knowledge on the topic.
Under this case, multimodal clarifying questions help enhance user performance, as they offer a richer context and enable users to respond more effectively with accurate answers. 
While answers to both multimodal and unimodal clarifying questions are textual, we hypothesize that answers provided in a multimodal setting are more valuable, due to the richer semantic information conveyed by the inclusion of images, ultimately resulting in improved document retrieval performance.

\begin{figure}[!t]
    \centering    
    \includegraphics[width=0.9\linewidth,clip]{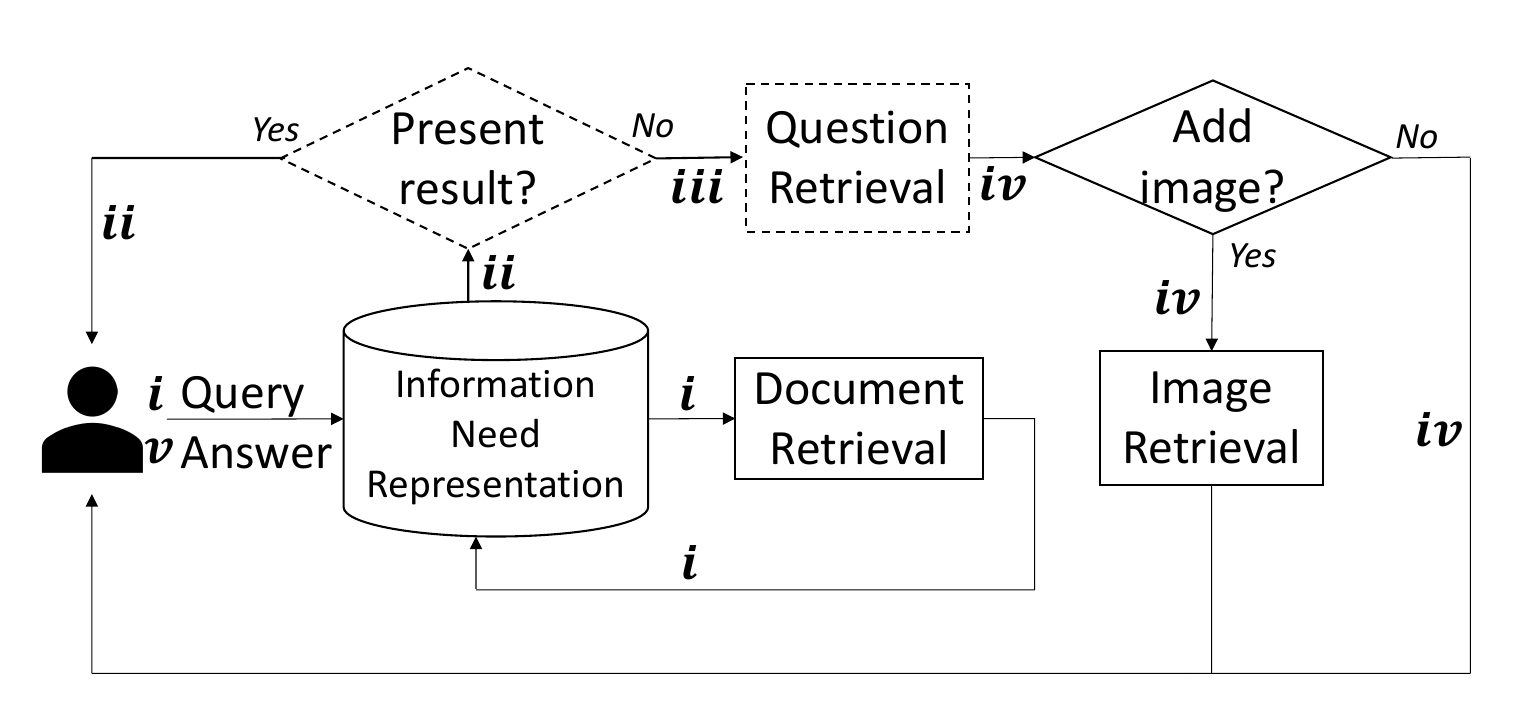}
    \caption{A workflow of adding MQC phase in a conversational search system. Hashed modules remain the same as in the unimodal clarification system presented in ~\cite{AliannejadiSigir19}.}
    \label{fig:workflow}
    \vspace{-1mm}
\end{figure}

Inspired by \citep{AliannejadiSigir19}, we propose a workflow for \ac{MQC} as shown in Fig.~\ref{fig:workflow}. 
The whole process begins when 
\begin{enumerate*}[label=(\roman*)]
\item a user submits a query, the system then predicts the information need by converting the query into an information need representation, and passes it to the document retrieval module to retrieve a ranked list of documents; 
\item at this point, the system decides whether to present the result to the user or ask a clarifying question, based on the confidence score assigned to the retrieved documents -- i.e., if the query is ambiguous or the information need is not clear, the confidence score falls below a threshold; 
\item in such cases, the system selects or generates a clarifying question to ask the user; 
\item after that, the system determines whether the question requires an image or not; if an image is needed, the image selection module returns an image to be added to the question. Otherwise, the question is presented to the user without any images; 
\item according to the clarifying questions and the corresponding images (if any), the user provides an answer. 
\end{enumerate*}
The system repeats this procedure until it reaches a high confidence score or the maximum number of questions are reached, and finally presents the retrieved documents to the user.

\vspace*{-2mm}
\section{Data Collection \& Analysis}
To facilitate MQC research, we describe how we create the dataset \OurData. We then perform analyses on \OurData to answer \textbf{RQ4}\footnote{The ethics statement of \OurData is listed in Appendix \ref{ethics}.}. 

\vspace*{-2mm}
\subsection{Data collection}
\label{sec:data-collection}
Our approach to constructing \OurData draws inspiration from existing unimodal clarification datasets~\cite{AliannejadiSigir19,aliannejadi2021building,Aliannejadi2020ConvAI3GC}. In \OurData, initial user queries can be subdivided into multiple facets that reflect the user's real intention. 
A set of clarifying questions are associated with every facet of the query. Afterwards, user responses are collected to answer each clarifying question and provide insight into the corresponding facet.
We first leverage topics from the TREC Web Track 2009--2012~\cite{Clarke2009OverviewOT} as user queries and reuse the collected subtopics as facets. To create a more enriched and diverse dataset that combines the existing resources with new instances, we utilize the clarifying questions from the text-only ClariQ dataset~\cite{Aliannejadi2020ConvAI3GC} while also collecting new ones from scratch. We then enrich all the clarifying questions with images and add  new answers to enhance the dataset's utility.

\header{Data collection pipeline}
We implement our data collection pipeline in 3 phases. 
\begin{enumerate*}[label=(\textbf{Phase \arabic*})]
    \item collect clarifying questions tailored to be multimodal from both ClariQ and from scratch;
    \item collect a diverse set of images that can be attached to the collected clarifying questions; and
    \item collect new answers for the clarifying questions presented with their corresponding images.
\end{enumerate*}

\header{Phase 1: Collecting multimodal clarifying questions}
The collection of suitable clarifying questions, specifically those that pertain to image attachment, is a crucial step in creating a high-quality MQC dataset. We gather these questions from two distinct sources:
\begin{enumerate*}[label=(\roman*)]
    \item existing ClariQ questions; and
    \item newly collected clarifying questions.
\end{enumerate*}
To obtain the first set of questions (which we call \textbf{set 1 questions}), we employ two expert annotators to classify the existing ClariQ questions as either \textbf{multimodal} (i.e., questions suitable for image attachments) or \textbf{unimodal} (i.e., questions not appropriate for image attachments).\footnote{We recruit the annotators from Appen (\url{https://appen.com/}).}  The annotators initially agree on 95\% of the annotated questions with Cohen's inter-annotators' agreement indicating a strong level of agreement ($\kappa = 0.82$). In case of disagreement, we resolve the conflict by asking the annotators to discuss and reach a final decision.
For the second set of questions~(\textbf{set 2 questions}), we design a \ac{HIT} on \ac{AMT}\footnote{\url{https://mturk.com}} where we ask annotators to generate new multimodal questions given the user query and ensure that the user facet remains undisclosed to them throughout the process. We provide detailed instructions on how to generate these questions and urge our annotators to follow these steps to create high-quality multimodal clarifying questions:
\begin{enumerate*}[label=(\roman*)]
    \item enter the user query into an image search engine (e.g., Google image search) and scan the first three pages; 
    \item scan the image-oriented question suggestions at the top of the result page, which provide useful hints for image-oriented aspects of the query;
    \item check the query auto-complete suggestions for additional hints; and
    \item write three questions, focusing on  different facets of the query.
\end{enumerate*}

\header{Phase 2: Image collection/attachment}
In Phase 2, we aim to associate each clarifying question obtained in Phase 1 with relevant images. To achieve this, we instruct expert annotators to use a search engine of their choice and find images that meet the requirements specified for each clarifying question. For instance, in the case of Fig.~\ref{fig:exmaple}, the annotator is guided to search for ``bike diagram'' in the image search engine, using both the query and clarifying question as a reference. Annotators are then required to select up to three images that they find most relevant to the clarifying question. To ensure diversity, we require that each selected image depicts a distinct aspect related to the query-clarifying question combination, and that annotators record the corresponding aspect of each image. E.g., annotators may provide the images of different types of bikes.


\header{Phase 3: Answer collection}
In Phase 3, we create a new AMT \ac{HIT} to collect answers for the multimodal clarifying questions. To ensure reliability of our results, we gather new answers for all questions instead of relying on the existing answer set from ClariQ. Our hypothesis is that the inclusion of images can have a significant impact on users' behavior, leading to inaccuracies in the text-only ClariQ answers. We provide the annotators with the original user query, topic facet, and multimodal clarifying questions along with their corresponding images.
We instruct them to act as users being part of an ongoing conversation with the facet as their real intention. We encourage them to provide natural language responses as if they were engaging in a dialogue with the system. We emphasize the importance of considering both the question and accompanying images when providing  answers. Our goal is to obtain accurate and informative answers that would help improve the performance of the \ac{MQC} system. We also adopt several quality control methods to ensure the quality of \OurData; cf.\ Appendix \ref{qualitycontrol}.


\begin{table}[]
    \caption{Statistics of the Melon dataset.}
    \label{tab:basic}
    \centering
    \begin{tabular}{lr@{}l}
    \toprule
    \# topics  & 298 \\
    \# facets &  1,070 \\
    \# all questions  &  4,969 \\
    \# set 1 questions & 3,365 \\
    \# set 2 questions & 1,604 \\
    Avg.~question per topic (std.) & 16&.67 (3.59) \\
    Avg.~\# terms per question (std.): & 9&.85 (2.63) \\
    \# images & 14,869 \\
    Avg.~\# images per question (std.) & 2&.99 (0.12)\\
    \# answers & 18,533 \\
    Avg.~\# answers per question (std.) & 3&.73 (1.61)\\    
    \bottomrule
    \end{tabular}
\end{table}

\vspace*{-2mm}
\subsection{Data analysis}
\label{sec:dataanalysis}

\header{Dataset statistics}
Table \ref{tab:basic} provides an overview of the basic statistics of \OurData. The dataset comprises a total of 298 search topics and 1070 facets. It consists of 4,969 multimodal clarifying questions accompanied by 14,869 associated images, resulting in an average of 2.99 images per question.  Among these questions, 67.7\% originate from ClariQ, whereas 32.3\% are newly collected. 
Additionally, the dataset includes 18,533 answers, with an average of 3.73 answers per question, equating to one answer per facet-question pair. To have a more comprehensive understanding of \OurData, we then compare the language usage in ClariQ questions~(\textbf{set 1 questions}) and \ac{AMT} questions~(\textbf{set 2 questions}). 
Due to page limitations, details are included in Appendix~\ref{termcompare}.

\begin{figure}
    \centering
    \begin{subfigure}{0.49\columnwidth}
        \includegraphics[width=\linewidth]{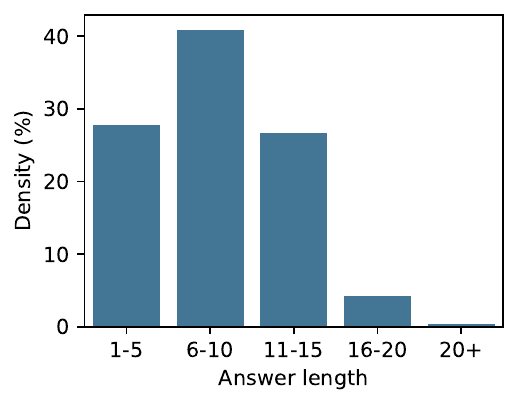}
        \vspace*{-7mm}
        \caption{Unimodal answers}
        \label{fig:answerdistribution-uni}
    \end{subfigure}
    \begin{subfigure}{0.49\columnwidth}
        \includegraphics[width=\linewidth]{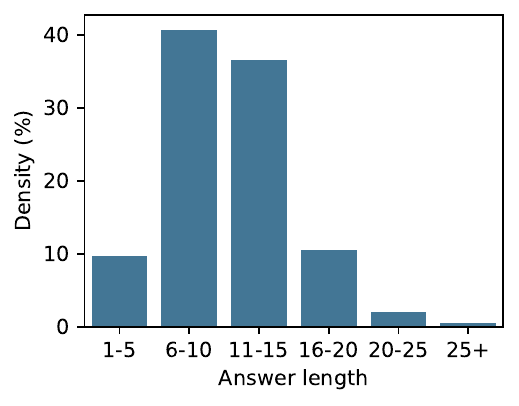}
        \vspace*{-7mm}
        \caption{Multimodal answers}
        \label{fig:answerdistribution-multi}
    \end{subfigure}   
    \caption{Distribution of answer length w.r.t terms in (a) unimodal and (b) multimodal datasets. Density represents the proportion of each type of answer in the answer set.}
    \label{fig:answerdistribution}
\end{figure}

\header{Answer characteristics}
\begin{table}[]
    \caption{Answer comparison between \OurData and ClariQ.}
    \label{tab:answeranalysis}
    \centering
    \begin{tabular}{l ccccc}
    \toprule
     &  Avg.  &  Mid. & Max. & Yes/no & Vocab.  \\
     &  terms (std.)  &  terms & terms & answers (\%) & size  \\
     \midrule
       ClariQ & \phantom{0}8.12 (4.58) & 9 & 30 & 10.25 & 4,561 \\
       \OurData & 10.76 (4.73) & 10 & 96 & \phantom{0}3.06 & 8,622 \\
    \bottomrule
    \end{tabular}
\end{table}
To address \textbf{RQ4} \emph{from a dataset perspective}, we analyze the differences in user responses between unimodal and multimodal conversational systems. We compare the user responses collected in \textbf{Phase 3} with the original responses of ClariQ --- responses to unimodal questions. 
Table~\ref{tab:answeranalysis} lists various characteristics of the answer collection of ClariQ and \OurData. Additionally, Figure~\ref{fig:answerdistribution} illustrates the distribution of answer lengths in the two datasets. 
When examining Table~\ref{tab:answeranalysis} and Figure~\ref{fig:answerdistribution}, we observe notable differences between the two datasets. The responses in \OurData exhibit more engagement, resulting in longer answers (with an average of 10.76 terms) compared to ClariQ (with an average of 8.12 terms). Furthermore, there is a lower percentage of one-word answers such as yes/no (3\% in \OurData vs. 10\% in ClariQ), indicating that the presence of multimodal clarifying questions encourages users to provide more detailed and informative responses.
Moreover, the vocabulary size of \OurData is nearly twice as large as that of ClariQ. This suggests that the inclusion of multimodal clarifying questions enables users to provide responses with richer semantic information, resulting in a more diverse and expanded vocabulary.

\begin{figure*}[htbp]
	\centering
 \vspace{-2mm}
	\begin{minipage}{0.39\linewidth}
		\centering
		\includegraphics[width=0.94\linewidth,height=45mm]{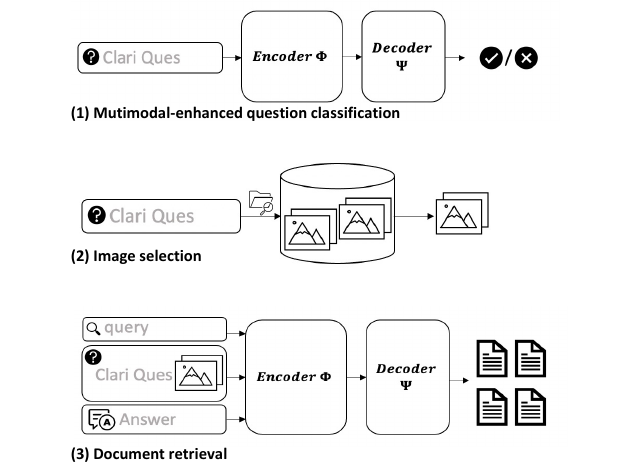}
		\caption{The three modules in the \OurModel model.}
		\label{fig:module}
	\end{minipage}
	\begin{minipage}{0.59\linewidth}
		\centering
		\includegraphics[width=\linewidth,height=45mm]{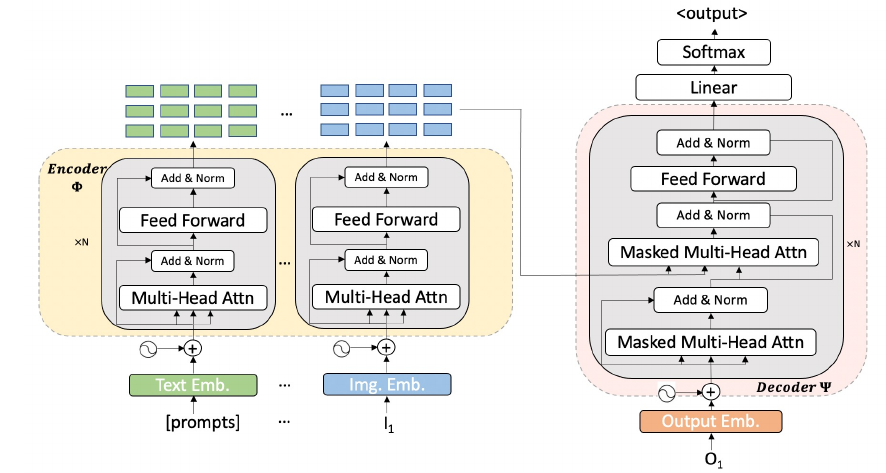}
		\caption{The detailed structure of Encoder $\Phi$ and Decoder $\Psi$.}
		\label{fig:model}
	\end{minipage}
\end{figure*}

\header{Summary}
We show that text-only clarifying questions can be enriched with images to create a multimodal dataset. We propose \OurData for offline evaluation of \ac{MQC} models.
Our analysis on the ClariQ and \OurData answers suggests that the inclusion of images in clarifying questions yields more comprehensive user responses; see Section~\ref{sec:results} for more on this in the context of \textbf{RQ4}. 

\vspace*{-3mm}
\section{Multimodal Query Clarification}
\label{model}

\header{Problem formulation}
Similar to Qulac~\cite{AliannejadiSigir19} and ClariQ~\cite{Aliannejadi2020ConvAI3GC}, given a list of topics denoted as $T=\{t_1,t_2,\ldots,t_k\}$, each topic is associated with a set of facets $\Gamma	=\{F_1,F_2,\ldots,F_k\}$, where $F_i=\{f_i^1,f_i^2,\ldots,f_i^{n_i}\}$ represents  the corresponding facets to the $i$-th topic $t_i$. $n_i$ denotes the number of facets of topic $t_i$. In addition, $Q_i=\{q_i^1,q_i^2,\ldots,q_i^{l_i}\}$ denotes the set of $l_i$ clarifying questions that  belong to topic $t_i$. Our approach is different from Qulac in the sense that each clarifying question $q_i^j$ is associated with a tag $b_i^j\in\{0,1\}$ representing if the question is multimodal, i.e., has images, or not. If $b_i^j = 1$, $I_i^j$ is the image set of question $q_i^j$. Furthermore, for each topic $t$, facet $f$, question $q$, and the associated question images $I$ (if any), we define an answer function $A(t,f,q,I)\rightarrow{a}$, which maps the current conversation context to a user answer. Following~\cite{aliannejadi2021building}, we borrow all the topic set $T$, facet set $\Gamma$, and relevance assessments from TREC Web Track 2009--2012. As described in Section~\ref{sec:data-collection}, we manually collect the question set $Q$ (\textbf{Phase 1}), the image set for each question $I$ (\textbf{Phase 2}), and the answers $A$ (\textbf{Phase 3}). 

\header{\OurModel architecture}
Fig.~\ref{fig:module} and~\ref{fig:model} illustrate the overall architecture of the model we propose for MQC task, \OurModel. 
Our approach builds on the workflow described in Section~\ref{sec:multimodal-clarificaiton} and consists of several modules.
Following the workflow in Fig. \ref{fig:workflow}, when a conversational system enters the question clarification phase after a user query submission (phase i, ii), a clarification question is retrieved (\textbf{clarification question retrieval module}). 
In phase iii, a question classification module is employed to judge if the question is multimodal (\textbf{multimodal-enhanced question classification module}). If yes, several images are selected (\textbf{image selection module}) in phase iv. In phase v, the users provide the response before all the additional information is fed into the document retrieval module to retrieve documents (\textbf{document retrieval module}). In \OurModel, we focus on the last three modules shown in Fig. \ref{fig:module}, as the other modules can be adopted from existing unimodal clarification systems~\cite{AliannejadiSigir19}.
As depicted in Fig.~\ref{model}, \OurModel is based on a multimodal generation model named VLT5~\cite{DBLP:conf/icml/ChoLTB21}. 
We develop a multitask fine-tuning strategy and use a single model with different prompts to fine-tune different subtasks~\cite{Lester2021ThePO,OpenAI2023GPT4TR}. 
The subtasks are detailed below.  

\header{Multimodal-enhanced question classification}
While all questions in the Melon dataset are assumed as multimodal by human annotators, our preliminary experiments have shown that attaching images to some questions can introduce additional noise and adversely affect the retrieval performance. 
We denote these questions as the text-enhanced questions (\textbf{\NEG}). However, for some other questions, adding images has a positive effect from the model's perspective. We denote them as the visually-enhanced questions (\textbf{\POS}). To classify questions as either TEQ or VEQ, multimodal-enhanced question classification involves training a binary classifier that labels the questions from the model prediction as suitable for image association or not. 
Section~\ref{sec:results} offers an analysis of the reasons behind the discrepancy between the predictions made by the model and those made by humans.

We train our classification model based on a multimodal generative model named VLT5~\cite{DBLP:conf/icml/ChoLTB21}. VLT5 is a state-of-art multimodal pretrained model that takes text and images as input and generates text as output. The text input of our model includes three parts: task prefix, user query, and clarifying question. The task prefix is a short prompt that differentiates this task from others. We use ``\verb|question classification:|'' as the task prefix. The user query and the clarifying questions are appended and separated by a special token [SEP]. All the text input is then tokenized and embedded before being passed to the encoder.  
Following~\cite{Raffel2019ExploringTL}, we incorporate relative position bias to each self-attention layer. As a result, the text input is represented as $e^t = \phi_T(\langle p_1\rangle,t,q)$, where $\phi_T$ represents the text-embedding function and $\langle p_1\rangle$ is the short prompt. 
Due to the lack of image input for this task, we mask the image domain of the model.  We train the model to generate the labels directly. For {\POS}s, it generates ``true'', and for {\NEG}s, it generates ``false.'' This can be represented as:
\vspace{-2mm}
\begin{equation}
    y_q = \operatorname{VLT5}(\phi_T(\langle p_1\rangle, t, q)),
    \vspace{-2mm}
\end{equation}
where $y_q$ is the label of each question, $t$,$q$ represent the topic and question. 

\header{Weakly supervised label generation}
We propose a weakly supervised label generation method to automatically generate ground-truth \POS/\NEG labels. We train two document retrieval models, where the first one takes a text-only (unimodal) input (e.g., BERT), called \TOR, and the second is a multimodal version (e.g., VisualBERT), called \MUR. To make a fair comparison, \TOR and \MUR share the same basic model structure and text input, including user query, clarifying question, and corresponding answer, each separated by the [SEP] token. For \MUR, we additionally attach the corresponding images of the clarifying questions as the image input. We compare the retrieval performance of these two models and calculate the relative improvement $\Delta{\operatorname{nDCG}} = \operatorname{nDCG}(\text{\MUR}) - \operatorname{nDCG}(\text{\TOR})$. If 
\MUR performs better ($\Delta{\operatorname{nDCG}}>0$), we assign a positive label to the clarifying question (\POS). 
Otherwise, the question is labeled as negative (\NEG). 
In our experiments, we use the average of $nDCG@\{1,3,5\}$ as the nDCG score.

\header{Image selection}
This module aims to rank the most relevant images given a clarifying question. 
We propose a simple but effective cross-modal retrieval method that does not require training. Inspired by the success of pre-trained image-text models in the cross-modal retrieval task, we adopt the pre-trained CLIP~\cite{Radford2021LearningTV} embedding to obtain the embedding of each clarifying question and image. We then rank the images based on their similarity scores to the question.
The selected image $I$ for question $q$ given candidate image set $I_q$ can be represented as:
\begin{equation} 
     I= \arg\max_{i\in I_q}(\cos(\operatorname{CLIP}(q), \operatorname{CLIP}(i)))~.
\end{equation}
We make our model choice based on preliminary experiments where we find that despite its simplicity, it shows good performance.

\header{Document retrieval}
Our document retrieval module ranks the documents for a topic, a given user query, a \POS clarifying question with images and the user answer. For the {\NEG}s, we keep the same model structure but mask the image input. 
Our model adopts the idea of generative retrieval method~\cite{DeCao2020AutoregressiveER,Chen2022CorpusBrainPA,Zhou2022DynamicRetrieverAP}, which has shown great research potential in the text-only retrieval domain under the recent success of generative NLP.

Similar to the multimodal-enhanced question classification module, this module utilizes the VLT5 model~\cite{DBLP:conf/icml/ChoLTB21} for the document retrieval task. However, we employ a different prompt specifically for document retrieval, using ``\verb|document retrieval:|'' as the task prefix. 
The rest of the text input contains the topic, the user query, the clarifying question, and the user answer, each separated by the [SEP] token. The tokenization and embedding methods are the same as the multimodal-enhanced question classification module. To obtain the embedding of image $I$, following previous works~\cite{Li2020OscarOA,Li2019VisualBERTAS,DBLP:conf/icml/ChoLTB21}, we first detect several object regions denoted as \ac{ROI}. All \acp{ROI} are detected using the object detection model FasterRCNN~\cite{Ren2015FasterRT} pre-trained on the Visual Genome dataset ~\cite{Krishna2016VisualGC}. We then add the \ac{ROI} features with \ac{ROI} bounding box coordinates and region ids $\in \{1,2,\ldots,n\} $ before fed into a linear layer. The final visual vector is represented as $e^v = \phi_V(I)$.

In the generative \ac{IR} literature, the model directly generates identifiers of a set of contexts given an input query~\cite{Chen2022CorpusBrainPA,DeCao2020AutoregressiveER,Chen2023AUG}. Following this approach, after feeding the image and text embedding into the VLT5 encoder, we aim to generate the sequence of identifiers of relevant documents. 
We extract the document keywords as the unique identifier of each document, for the reason that keyword represents the most important part of each document and has a natural language format. For each document, we take the top-5 keywords as the corresponding identifier. We train the decoder to generate the identifier  of the top-5 ground-truth relevant pages, each concatenated by the [SEP] token. The training process can be represented as:
\vspace{-2mm}
\begin{equation} 
    y_d = \operatorname{VLT5}(\phi_T(\langle{t_2}\rangle, t, q, a), \phi_V(I))~,
    \vspace{-1mm}
\end{equation}
where $y_d$ is the generated document identifier sequence, $t_2$ is the prompt of the task, $a$ is the user answer.

For multimodal-enhanced classification and document retrieval, we adopt the standard generation loss when fine-tuning VLT5. During inference, we use constrained beam search~\cite{DeCao2020AutoregressiveER} to rank documents. This allows us to limit the generated content to be within the pre-defined candidate set with a generative score, i.e., the keyword identifiers of all documents in our corpus. 


\vspace*{-2mm}
\section{Experiments}
We experiment on the \OurData dataset by comparing \OurModel with several state-of-the-art methods, including lexical methods (i.e., OriginalQuery, QL, BM25, LambdaMART), pipeline-based methods (i.e., BERT+CLIP+CLIP, BERT+CLIP+VLT5), methods under a multi-task framework (i.e., BERT, T5, VisualBERT, VisualBERT\textunderscore{w/o QC}), and variants of \OurModel (i.e., \OurModelNS\textunderscore{w/o QC},  
 \OurModelNS\textunderscore{random-image}, \OurModelNS\textunderscore{trel-image}, \OurModelNS\textunderscore{oracle-best-image}). 
Appendix \ref{setup} details our experimental setup and baselines.

\begin{table*}[]\small
    \caption{Experimental results of \OurModel compared with baselines. QC represents the multimodal-enhanced question classification module. Img. represents whether the model takes image as input or not. Numbers in bold represent the best-performing model. * denotes \OurModel shows significant improvement under the significance test. All the numbers are shown in \%.}
    \label{mainexperiment}
    \centering
    \setlength{\tabcolsep}{3mm}
    \begin{adjustbox}{max width=1\textwidth}
    \begin{tabular}{l ccccccccccc}
    \toprule

         & Img. & MRR & P@1 & P@3 & P@5   & nDCG@1 & nDCG@3 & nDCG@5  &ERR@1 &ERR@3 &ERR@5 \\
    \midrule
        
        OriginalQuery & \xmark & 14.06 &18.75 &14.58 &11.88  & \phantom{0}5.05 & \phantom{0}3.64 & \phantom{0}3.77  & \phantom{0}2.44 & \phantom{0}2.83 & \phantom{0}3.28 \\
        QL & \xmark & 14.71 & 17.64 & 19.61 & 20.00  & 12.35& 12.49 &12.09 & \phantom{0}4.96 & \phantom{0}6.41 & \phantom{0}8.08 \\
        BM25 & \xmark & 20.31 & 23.44 & 24.48 & 24.69 & 11.59 &10.07 & \phantom{0}9.98  & \phantom{0}4.59 & \phantom{0}7.05 & \phantom{0}7.19\\
        
        LambdaMART & \xmark & 24.39 & 23.95 & 25.41 & 24.78 & 12.32 &11.89 & 11.39 & 
        \phantom{0}4.70 & \phantom{0}9.02 & \phantom{0}9.62 \\
        \midrule
        BERT+CLIP+CLIP & \cmark & 35.62 & 29.85 & 28.87 & 27.69 & 17.02 & 18.17 & 17.19 & 6.73 & 12.39 & 15.34 \\
        BERT+CLIP+VLT5 &  \cmark & 37.90 & 34.52 & 33.70 & 32.85 & 20.92 & 21.15 & 19.09 & 9.26 &13.48 &13.54 \\
        \midrule
        BERT & \xmark & 26.43 & 25.00 & 26.56 &28.24 & 13.17 & 13.70  &14.52 & \phantom{0}5.76 & 9.16 & 10.77\\
        T5 & \xmark & 27.56 & 25.10 & 26.04 & 26.88 & 13.62 & 14.16 &14.28  & \phantom{0}5.18 &9.84 &10.70 \\
        VisualBERT& \cmark & 41.56 & 33.75 & 32.01 & 31.94 &19.55 & 20.80 &18.85  & \phantom{0}7.99 &15.13 &14.39 \\
        VisualBERT\textunderscore{w/o QC} & \cmark & 33.28 & 30.54 & 32.31 & 32.45 & 15.31 &16.16 &19.55  & \phantom{0}6.88& 12.47 & 12.85  \\ 
        \midrule
        \OurModel & \cmark & \textbf{54.70}\rlap{*} & \textbf{53.38}\rlap{*} & \textbf{40.47}\rlap{*} & \textbf{36.65}\rlap{*}  & \textbf{30.66}\rlap{*} & \textbf{24.57}\rlap{*} &\textbf{23.81}\rlap{*}  & \textbf{15.63}\rlap{*} & \textbf{20.21}\rlap{*}&  \textbf{21.64}\rlap{*} \\
        \OurModelNS\textunderscore{w/o QC}& \cmark   & 46.32 & 40.24 & 36.65 & 30.97 & 20.50 & 18.26 &20.32  & 10.24 & 14.57 & 15.90  \\     
        \OurModelNS\textunderscore{trel-image}& \cmark&51.64& 50.28 & 38.01& 34.60 & 27.15 & 22.66 & 21.84 & 12.77& 14.80 & 19.06  \\
        \OurModelNS\textunderscore{random-image}& \cmark  & 35.50 & 33.85 & 31.70 & 29.20  & 18.76 & 18.91 &19.15  & \phantom{0}5.67 &12.63 & 13.40    \\
        \OurModelNS\textunderscore{oracle-best-image}& \cmark & 62.50 & 60.12 & 44.20 & 41.21 & 38.65 & 28.91 &25.03  & 19.78 & 23.81& 24.35 \\
        \bottomrule
    \end{tabular}
    \end{adjustbox}
\end{table*}

\vspace*{-2mm}
\subsection{Results \& Analyses}
\label{sec:results}
\header{Primary findings} In Table~\ref{mainexperiment} we report the performance of \OurModel and other several baselines. We also perform some preliminary experiments on the design of \OurModel and list the results in Appendix \ref{intermediate} and \ref{modelstructure}. 
Our findings indicate that incorporating clarifying questions significantly improves the performance of document retrieval compared to directly using the query alone (i.e., OriginalQuery) consistent with findings in~\citep{AliannejadiSigir19}. Moreover, large pre-trained language models like BERT demonstrate superior performance compared to lexical methods such as BM25. Furthermore, multi-task methods with a more compact structure outperform pipeline-based methods (e.g., BERT+CLIP+VLT5), allowing for the sharing of feature representations across tasks. Detailed analyses are listed below.

\begin{figure}[h]
    \centering
    \begin{subfigure}{0.49\columnwidth}
        \includegraphics[width=\linewidth]{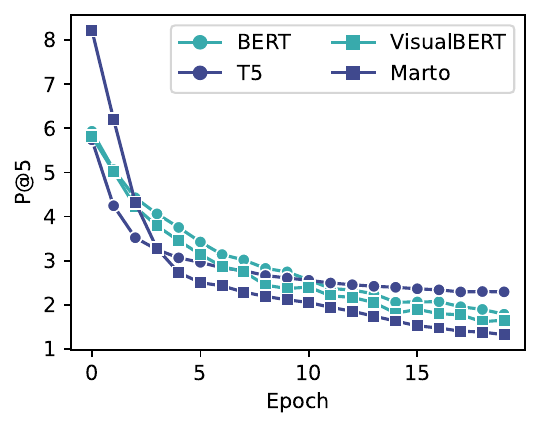}
        \vspace*{-7mm}
        \caption{Loss per epoch}
        \label{fig:training-loss}
    \end{subfigure}
    \begin{subfigure}{0.49\columnwidth}
        \includegraphics[height=33mm]{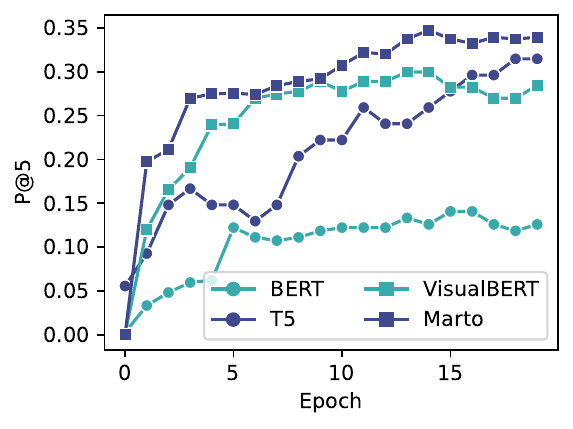}
        \vspace*{-7mm}
        \caption{P@5 per epoch}
        \label{fig:training-p5}
    \end{subfigure}
    \caption{Training loss (a) and validation P@5 (b) score at different training epochs.}
    \label{fig:training}
    \vspace{-2.5mm}
\end{figure}

\header{Image vs. no image (RQ1)}
Regarding the impact of images on the document retrieval task (as addressed above), we observe that clarification systems with multimodal information outperform those relying solely on text  (e.g., BERT vs.\ VisualBERT) in Table~\ref{mainexperiment}. Also, we find that adding images can lead to up to 90\% improvements in performance (\OurModel vs. T5). In terms of the convergence, as shown in Fig.~\ref{fig:training},  multimodal methods (square marker) take less time to reach good performance than text-only methods (circle marker), demonstrating the positive impact of incorporating multimodal information on model optimization and training time. Also, classifying the questions into \POS and \NEG categories before directly performing the retrieval task can further boost the results. For VisualBERT, removing the multimodal-enhanced question classification module results in a 3.21\% decrease in P@1 (VisualBERT vs.\ VisualBERT\textunderscore{w/o QC}), while for \OurModel, P@1 is reduced by 13.14\% (\OurModel vs.\ \OurModelNS\textunderscore{w/o QC}). We also record the percentage of topics and questions that benefit from the addition of images. We find that the retrieval performance of nearly 80\% of questions gets improved by adding the relevant images, covering around 90\% topics. This finding confirms that while the majority of questions benefit from image attachment at the model level, some do not. Thus, predicting the suitability of questions for image attachment is necessary.

\header{Impact of image variations (RQ2)}
To further investigate the impact of images, we adopt several variants: \OurModelNS\textunderscore{random-image}, \OurModelNS\textunderscore{trel-image}, and \OurModelNS\textunderscore{oracle-best-image}. We observe that even including a random image helps improve the retrieval process. Our findings further indicate that selecting and attaching the right image has a strong impact on the performance. Compared with attaching the random image (\OurModelNS\textunderscore{random-image}), choosing the topic related image (\OurModelNS\textunderscore{trel-image}) demonstrates a substantial improvement in performance, with a 45\% increase in MRR rate. Overall, using the CLIP model for selecting question-relevant images in \OurModel shows further advantage, verifying the effectiveness of our image selection module.
 Overall, \OurModel demonstrates superior performance compared to another multimodal models and variants across multiple evaluation metrics. 


\begin{table}[t]
    \caption{Comparison of the memory usage, the number of model parameters, and training inference time per epoch.}
    \label{tab:training}
    \centering
     \small
    \begin{tabular}{l cccc}
    \toprule
        Model & Memory & Parameters & Train. time & Inf. time \\
    \midrule
        BERT & 5621M & 110M & 7.30 min & 0.82 min \\
        T5 & 8753M & 220M & 0.48 min & 0.33 min  \\
        VisualBERT & 7189M & 110M & 8.26 min &1.13 min \\
    \midrule
        \OurModel & 9965M &  220M & 0.77 min & 0.67 min \\
    \bottomrule
    \end{tabular}
\end{table}

\begin{table}[h]
    \caption{Comparison of \OurModel's performance under \OurData subset where all the questions originate from ClariQ. }
    \label{datasetanalysis}
    \centering
    \small
    \setlength{\tabcolsep}{1.3mm}
    \begin{tabular}{lc ccccc}
    \toprule
         Ans src. & Img. & MRR & P@1 & P@5 & nDCG@1 & nDCG@5  \\
    \midrule
        ClariQ & \xmark & 30.95 & 28.57 & 29.52 & 14.40 & 14.19  \\
        ClariQ & \cmark & 39.77 & 36.46 & 32.50 & 20.73 & 18.16  \\
        \OurData & \xmark & 28.85 & 30.95 & 28.10 & 19.00 & 15.79  \\
        \OurData & \cmark & \textbf{54.81}\rlap{*} & \textbf{53.73}\rlap{*} & \textbf{40.18}\rlap{*} & \textbf{30.98}\rlap{*} & \textbf{23.90}\rlap{*} \\
    \bottomrule
    \end{tabular}
    \vspace{-1mm}
\end{table}
\begin{table*}[!h]
\small
    \centering
    \caption{Case study on the human-labeled multimodal questions. The up arrow means performance increases after adding images (\POS), while the down arrow means performance decreases after adding images (\NEG).}
    \setlength{\tabcolsep}{1mm}
    \begin{tabular}{c p{1.5cm} p{1.5cm} p{4.0cm} p{3.5cm} p{1.8cm} p{2.5cm} c}
    \toprule
        \textbf{Idx} & \textbf{Category} & \textbf{Topic} & \textbf{Facet} & \multicolumn{2}{c}{\textbf{Clarification question}} & \textbf{Answer} & \textbf{Pred}   \\
        \midrule
        1 & Shopping & bowflex power pro & \raggedright Find information about the Bowflex Power Pro. & Do you want to buy some parts for this equipment? & \raisebox{-.7\height}{\includegraphics[width=1.5cm,height=0.8cm]{sections/Figures/case1.pdf}} & Yes, I'm interested in what the material is.& $\uparrow$\\
          \midrule
        2 & Location-related & map of Brazil & \raggedright I am looking for information about taking a vacation trip to Brazil. & Would you like to see a specific region of Brazil? & \raisebox{-.7\height}{\includegraphics[width=1.5cm,height=1cm]{sections/Figures/case2.pdf}} & Yes. The pyramid in the picture looks interesting. & $\uparrow$ \\
        \midrule
        3 & General & titan & Find the homepage for Titan motorcycles. & Are you interested in a specific titan? & \raisebox{-.7\height}{\includegraphics[width=1.5cm,height=0.9cm]{sections/Figures/case3.pdf}} & No, I'm not interested in the anime, but the motorcycle. & $\uparrow$ \\
        \midrule
        4 & Recipe-related & salads & \raggedright Find salads that are both nutritious and vegetarian-friendly. & Would you like to know about the recipe of this? & \raisebox{-.7\height}{\includegraphics[width=1.5cm,height=0.8cm]{sections/Figures/case4.pdf}} & Yes, it looks good. & $\downarrow$\\
        \midrule
        5 & Categorical & insects & \raggedright  Find information on different types of insects.&Here's one example, is that what you need? & \raisebox{-.7\height}{\includegraphics[width=1.5cm,height=0.9cm]{sections/Figures/case5.pdf}}& Yes, exactly. & $\downarrow$\\
        \bottomrule
    \end{tabular}    
    \label{tab:casestudy}
\end{table*}

\header{Generative vs.~discriminative modeling (RQ3)}
Table \ref{mainexperiment} shows that in both unimodal and multimodal scenarios, compared with discriminative retrieval models (e.g., BERT, VisualBERT), generative retrieval methods (e.g., T5, \OurModel) improve overall performance. This indicates that generative models are more effective in incorporating and unifying multimodal (image + text) information. To further address \textbf{RQ3} on the efficiency of generative retrieval models, we conduct experiments and analyze the training progress in terms of loss and validation P@5 score (Fig.~\ref{fig:training}). We compare the efficiency of these models and report the training and inference time of each epoch in Table~\ref{tab:training}.
As shown in Fig.~\ref{fig:training}, we observe that:
\begin{enumerate*}[label=(\roman*)]
    \item generative models (dark blue) outperform discriminative models (light blue) in both performance and training efficiency. Despite having more parameters than VisualBERT, \OurModel requires significantly less training time (0.77 vs. 8.26) and inference time (0.67 vs. 1.13).
    \item \OurModel achieves the best overall convergence speed and the best validation score with the shortest time among all baselines, indicating that it can easily adapt to the downstream retrieval task with the same learning objective as in the pre-training stage.
\end{enumerate*}

\header{Multimodal impact on user answers (RQ4)} 
To answer RQ4 \emph{from the model perspective}, we report the document retrieval performance on a subset of our dataset. We focus on \OurData questions sourced from ClariQ (\textbf{set 1 questions}) and generate different variants by pairing them with various answer sources (see Table~\ref{datasetanalysis}).  \textit{A src.}~column specifies where the user's answer originates from. The outcomes indicate a significant performance improvement by replacing the ClariQ answers with \OurData answers (2nd row vs.\ 4th row). This indicates that users can offer more comprehensive answers when provided with multimodal questions, as the answers derived from \OurData are closely aligned with the corresponding images. These results highlight the advantages of collecting new answers, aligning with the findings in Section~\ref{sec:dataanalysis}. Significantly, we observe that multimodal information improves the results both in ClariQ and \OurData answer sets (1st row vs.~2nd row; 3rd row vs.~4th row), suggesting that the images serve as a highly informative resource for addressing the underlying user intent, aligning with 
findings for \textbf{RQ1}.

\begin{figure}[h]
    \centering
    \includegraphics[width=0.8\columnwidth]{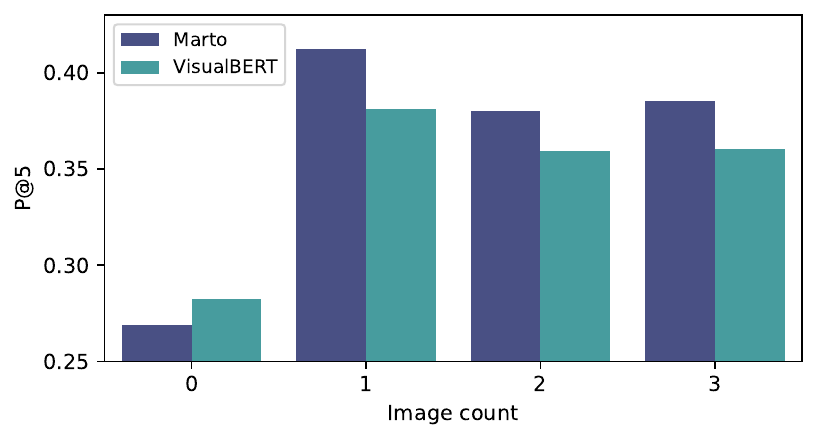}
    \caption{The oracle P@5 retrieval performance concerning the number of images attached to the question.}
    \label{fig:imgquantity}
\end{figure}

\header{Impact of the number of images}
In order to assess the impact of the number of attached images on the performance of \OurModel and VisualBERT, we compare their P@5 performance, as depicted in Fig.~\ref{fig:imgquantity}.
Our observations reveal that incorporating images into the clarifying questions enhances the performance of both models compared to using text-only questions (image count 0). Furthermore, we find that attaching the top-one image yields the best performance, surpassing the results obtained when attaching 2 or 3 images.
These findings validate the effectiveness of the image selection module in our model, emphasizing the importance of selecting the most relevant image to enhance overall performance.

\header{Case study}
To illustrate the correlation between human judgment and model prediction, we present a set of human-assessed multimodal questions and provide the corresponding model's prediction; see Table~\ref{tab:casestudy}.
In most cases, including images in clarifying questions can provide valuable extra information. Notably, in case 3, the clarifying question may be focused on a different aspect than the underlying user intent, while the negative feedback from users after viewing the image can still be valuable. However, there are some cases where images can be misleading. One such case is when the facets contain specific restrictions that are hard for the system to notice. For example, in case 4, the user intends to provide a clarifying question for a vegetarian-friendly recipe. Unfortunately, the image provided includes ingredients like shrimp that contradict the hidden facet of vegetarian-friendly recipes. Another case where images can introduce bias is in case 5,  where the system misinterprets the image as an indication of the user's specific interest in butterfly species. Consequently, it could prioritize retrieving information related only to butterflies rather than diverse insect species.


\vspace*{-2mm}
\section{Conclusion}
We investigate the novel task of asking multimodal query clarification in a text-dominant conversation.
To provide an offline evaluation method for this system, we collect a dataset named \OurData, with over 4k multimodal clarifying questions and 14k corresponding images. We also propose a multimodal query clarification model named \OurModel which adopts a prompt-based generative fine-tuning strategy for different subtasks. Experiments show that adding images can help improve retrieval results and lead to a significant performance lift. Additionally, multimodal contents result in more contextualized answers with richer semantic information. Compared to discriminative models, \OurModel demonstrates superiority in the document retrieval task.  
For future research, we list some ongoing directions in Appendix \ref{ongingdirections}.


\clearpage


\clearpage
\bibliographystyle{ACM-Reference-Format}
\balance
\bibliography{references}


\begin{thebibliography}{81}


\ifx \showCODEN    \undefined \def \showCODEN     #1{\unskip}     \fi
\ifx \showDOI      \undefined \def \showDOI       #1{#1}\fi
\ifx \showISBNx    \undefined \def \showISBNx     #1{\unskip}     \fi
\ifx \showISBNxiii \undefined \def \showISBNxiii  #1{\unskip}     \fi
\ifx \showISSN     \undefined \def \showISSN      #1{\unskip}     \fi
\ifx \showLCCN     \undefined \def \showLCCN      #1{\unskip}     \fi
\ifx \shownote     \undefined \def \shownote      #1{#1}          \fi
\ifx \showarticletitle \undefined \def \showarticletitle #1{#1}   \fi
\ifx \showURL      \undefined \def \showURL       {\relax}        \fi
\providecommand\bibfield[2]{#2}
\providecommand\bibinfo[2]{#2}
\providecommand\natexlab[1]{#1}
\providecommand\showeprint[2][]{arXiv:#2}

\bibitem[\protect\citeauthoryear{Aliannejadi, Azzopardi, Zamani, Kanoulas,
  Thomas, and Craswell}{Aliannejadi et~al\mbox{.}}{2021a}]%
        {Aliannejadi2021AnalysingMI}
\bibfield{author}{\bibinfo{person}{Mohammad Aliannejadi}, \bibinfo{person}{Leif
  Azzopardi}, \bibinfo{person}{Hamed Zamani}, \bibinfo{person}{Evangelos
  Kanoulas}, \bibinfo{person}{Paul Thomas}, {and} \bibinfo{person}{Nick
  Craswell}.} \bibinfo{year}{2021}\natexlab{a}.
\newblock \showarticletitle{Analysing Mixed Initiatives and Search Strategies
  during Conversational Search}. In \bibinfo{booktitle}{\emph{Proceedings of
  the 30th ACM International Conference on Information \& Knowledge
  Management}}.
\newblock


\bibitem[\protect\citeauthoryear{Aliannejadi, Kiseleva, Chuklin, Dalton, and
  Burtsev}{Aliannejadi et~al\mbox{.}}{2021b}]%
        {aliannejadi2021building}
\bibfield{author}{\bibinfo{person}{Mohammad Aliannejadi},
  \bibinfo{person}{Julia Kiseleva}, \bibinfo{person}{Aleksandr Chuklin},
  \bibinfo{person}{Jeff Dalton}, {and} \bibinfo{person}{Mikhail Burtsev}.}
  \bibinfo{year}{2021}\natexlab{b}.
\newblock \showarticletitle{Building and Evaluating Open-Domain Dialogue
  Corpora with Clarifying Questions}. In \bibinfo{booktitle}{\emph{{EMNLP}}}.
\newblock


\bibitem[\protect\citeauthoryear{Aliannejadi, Kiseleva, Chuklin, Dalton, and
  Burtsev}{Aliannejadi et~al\mbox{.}}{2020}]%
        {Aliannejadi2020ConvAI3GC}
\bibfield{author}{\bibinfo{person}{Mohammad Aliannejadi},
  \bibinfo{person}{Julia Kiseleva}, \bibinfo{person}{Aleksandr Chuklin},
  \bibinfo{person}{Jeffrey Dalton}, {and} \bibinfo{person}{Mikhail~S.
  Burtsev}.} \bibinfo{year}{2020}\natexlab{}.
\newblock \showarticletitle{ConvAI3: Generating Clarifying Questions for
  Open-Domain Dialogue Systems (ClariQ)}.
\newblock \bibinfo{journal}{\emph{ArXiv}}  \bibinfo{volume}{abs/2009.11352}
  (\bibinfo{year}{2020}).
\newblock


\bibitem[\protect\citeauthoryear{Aliannejadi, Zamani, Crestani, and
  Croft}{Aliannejadi et~al\mbox{.}}{2019}]%
        {AliannejadiSigir19}
\bibfield{author}{\bibinfo{person}{Mohammad Aliannejadi},
  \bibinfo{person}{Hamed Zamani}, \bibinfo{person}{Fabio Crestani}, {and}
  \bibinfo{person}{W.~Bruce Croft}.} \bibinfo{year}{2019}\natexlab{}.
\newblock \showarticletitle{Asking Clarifying Questions in Open-Domain
  Information-Seeking Conversations}. In \bibinfo{booktitle}{\emph{Proceedings
  of the 2019 International {ACM} {SIGIR} Conference on Research and
  Development in Information Retrieval}}.
\newblock


\bibitem[\protect\citeauthoryear{Allen, Guinn, and Horvtz}{Allen
  et~al\mbox{.}}{1999}]%
        {Allen1999MixedinitiativeI}
\bibfield{author}{\bibinfo{person}{James Allen}, \bibinfo{person}{Curry~I.
  Guinn}, {and} \bibinfo{person}{E. Horvtz}.} \bibinfo{year}{1999}\natexlab{}.
\newblock \showarticletitle{Mixed-initiative interaction}.
\newblock \bibinfo{journal}{\emph{IEEE Intelligent Systems \& Their
  Applications}}  \bibinfo{volume}{14} (\bibinfo{year}{1999}).
\newblock


\bibitem[\protect\citeauthoryear{Bevilacqua, Ottaviano, Lewis, tau Yih, Riedel,
  and Petroni}{Bevilacqua et~al\mbox{.}}{2022}]%
        {Bevilacqua2022AutoregressiveSE}
\bibfield{author}{\bibinfo{person}{Michele Bevilacqua},
  \bibinfo{person}{Giuseppe Ottaviano}, \bibinfo{person}{Patrick Lewis},
  \bibinfo{person}{Wen tau Yih}, \bibinfo{person}{Sebastian Riedel}, {and}
  \bibinfo{person}{Fabio Petroni}.} \bibinfo{year}{2022}\natexlab{}.
\newblock \showarticletitle{Autoregressive Search Engines: Generating
  Substrings as Document Identifiers}.
\newblock \bibinfo{journal}{\emph{ArXiv}}  \bibinfo{volume}{abs/2204.10628}
  (\bibinfo{year}{2022}).
\newblock


\bibitem[\protect\citeauthoryear{Bokhari and Hasan}{Bokhari and Hasan}{2013}]%
        {Bokhari2013MultimodalIR}
\bibfield{author}{\bibinfo{person}{Mohammad~Ubaidullah Bokhari} {and}
  \bibinfo{person}{Faraz Hasan}.} \bibinfo{year}{2013}\natexlab{}.
\newblock \showarticletitle{Multimodal Information Retrieval: Challenges and
  Future Trends}.
\newblock \bibinfo{journal}{\emph{International Journal of Computer
  Applications}}  \bibinfo{volume}{74} (\bibinfo{year}{2013}).
\newblock


\bibitem[\protect\citeauthoryear{Boni and Manandhar}{Boni and
  Manandhar}{2003}]%
        {Boni2003AnAO}
\bibfield{author}{\bibinfo{person}{Marco~De Boni} {and} \bibinfo{person}{Suresh
  Manandhar}.} \bibinfo{year}{2003}\natexlab{}.
\newblock \showarticletitle{An Analysis of Clarification Dialogue for Question
  Answering}. In \bibinfo{booktitle}{\emph{Proceedings of the 2003 Conference
  of North American Chapter of the Association for Computational Linguistics}}.
\newblock


\bibitem[\protect\citeauthoryear{Braslavski, Savenkov, Agichtein, and
  Dubatovka}{Braslavski et~al\mbox{.}}{2017}]%
        {Braslavski2017WhatDY}
\bibfield{author}{\bibinfo{person}{Pavel Braslavski}, \bibinfo{person}{Denis
  Savenkov}, \bibinfo{person}{Eugene Agichtein}, {and} \bibinfo{person}{Alina
  Dubatovka}.} \bibinfo{year}{2017}\natexlab{}.
\newblock \showarticletitle{What Do You Mean Exactly?: Analyzing Clarification
  Questions in CQA}. In \bibinfo{booktitle}{\emph{Proceedings of the 2017
  Conference on Conference Human Information Interaction and Retrieval}}.
\newblock


\bibitem[\protect\citeauthoryear{Brown, Mann, Ryder, Subbiah, Kaplan, Dhariwal,
  Neelakantan, Shyam, Sastry, Askell, Agarwal, Herbert-Voss, Krueger, Henighan,
  Child, Ramesh, Ziegler, Wu, Winter, Hesse, Chen, Sigler, Litwin, Gray, Chess,
  Clark, Berner, McCandlish, Radford, Sutskever, and Amodei}{Brown
  et~al\mbox{.}}{2020}]%
        {Brown2020LanguageMA}
\bibfield{author}{\bibinfo{person}{Tom~B. Brown}, \bibinfo{person}{Benjamin
  Mann}, \bibinfo{person}{Nick Ryder}, \bibinfo{person}{Melanie Subbiah},
  \bibinfo{person}{Jared Kaplan}, \bibinfo{person}{Prafulla Dhariwal},
  \bibinfo{person}{Arvind Neelakantan}, \bibinfo{person}{Pranav Shyam},
  \bibinfo{person}{Girish Sastry}, \bibinfo{person}{Amanda Askell},
  \bibinfo{person}{Sandhini Agarwal}, \bibinfo{person}{Ariel Herbert-Voss},
  \bibinfo{person}{Gretchen Krueger}, \bibinfo{person}{T.~J. Henighan},
  \bibinfo{person}{Rewon Child}, \bibinfo{person}{Aditya Ramesh},
  \bibinfo{person}{Daniel~M. Ziegler}, \bibinfo{person}{Jeff Wu},
  \bibinfo{person}{Clemens Winter}, \bibinfo{person}{Christopher Hesse},
  \bibinfo{person}{Mark Chen}, \bibinfo{person}{Eric Sigler},
  \bibinfo{person}{Mateusz Litwin}, \bibinfo{person}{Scott Gray},
  \bibinfo{person}{Benjamin Chess}, \bibinfo{person}{Jack Clark},
  \bibinfo{person}{Christopher Berner}, \bibinfo{person}{Sam McCandlish},
  \bibinfo{person}{Alec Radford}, \bibinfo{person}{Ilya Sutskever}, {and}
  \bibinfo{person}{Dario Amodei}.} \bibinfo{year}{2020}\natexlab{}.
\newblock \showarticletitle{Language Models are Few-Shot Learners}.
\newblock \bibinfo{journal}{\emph{ArXiv}}  \bibinfo{volume}{abs/2005.14165}
  (\bibinfo{year}{2020}).
\newblock


\bibitem[\protect\citeauthoryear{Cao, Izacard, Riedel, and Petroni}{Cao
  et~al\mbox{.}}{2020}]%
        {DeCao2020AutoregressiveER}
\bibfield{author}{\bibinfo{person}{Nicola~De Cao}, \bibinfo{person}{Gautier
  Izacard}, \bibinfo{person}{Sebastian Riedel}, {and} \bibinfo{person}{Fabio
  Petroni}.} \bibinfo{year}{2020}\natexlab{}.
\newblock \showarticletitle{Autoregressive Entity Retrieval}.
\newblock \bibinfo{journal}{\emph{ArXiv}}  \bibinfo{volume}{abs/2010.00904}
  (\bibinfo{year}{2020}).
\newblock


\bibitem[\protect\citeauthoryear{Chang, Narang, Suzuki, Cao, Gao, and
  Bisk}{Chang et~al\mbox{.}}{2021}]%
        {Chang2021WebQAMA}
\bibfield{author}{\bibinfo{person}{Yingshan Chang},
  \bibinfo{person}{Mridu~Baldevraj Narang}, \bibinfo{person}{Hisami Suzuki},
  \bibinfo{person}{Guihong Cao}, \bibinfo{person}{Jianfeng Gao}, {and}
  \bibinfo{person}{Yonatan Bisk}.} \bibinfo{year}{2021}\natexlab{}.
\newblock \showarticletitle{WebQA: Multihop and Multimodal QA}. In
  \bibinfo{booktitle}{\emph{Proceedings of the 2022 IEEE/CVF Conference on
  Computer Vision and Pattern Recognition}}.
\newblock


\bibitem[\protect\citeauthoryear{Chen, Zhang, Guo, de~Rijke, Liu, Fan, and
  Cheng}{Chen et~al\mbox{.}}{2023b}]%
        {Chen2023AUG}
\bibfield{author}{\bibinfo{person}{Jiangui Chen}, \bibinfo{person}{Ruqing
  Zhang}, \bibinfo{person}{Jiafeng Guo}, \bibinfo{person}{Maarten de Rijke},
  \bibinfo{person}{Yiqun Liu}, \bibinfo{person}{Yixing Fan}, {and}
  \bibinfo{person}{Xueqi Cheng}.} \bibinfo{year}{2023}\natexlab{b}.
\newblock \showarticletitle{A Unified Generative Retriever for
  Knowledge-Intensive Language Tasks via Prompt Learning}.
\newblock \bibinfo{journal}{\emph{ArXiv}}  \bibinfo{volume}{abs/2304.14856}
  (\bibinfo{year}{2023}).
\newblock


\bibitem[\protect\citeauthoryear{Chen, Zhang, Guo, Fan, and Cheng}{Chen
  et~al\mbox{.}}{2022a}]%
        {Chen2022GEREGE}
\bibfield{author}{\bibinfo{person}{Jiangui Chen}, \bibinfo{person}{Ruqing
  Zhang}, \bibinfo{person}{Jiafeng Guo}, \bibinfo{person}{Yixing Fan}, {and}
  \bibinfo{person}{Xueqi Cheng}.} \bibinfo{year}{2022}\natexlab{a}.
\newblock \showarticletitle{GERE: Generative Evidence Retrieval for Fact
  Verification}.
\newblock \bibinfo{journal}{\emph{Proceedings of the 45th International ACM
  SIGIR Conference on Research and Development in Information Retrieval}}
  (\bibinfo{year}{2022}).
\newblock


\bibitem[\protect\citeauthoryear{Chen, Zhang, Guo, Liu, Fan, and Cheng}{Chen
  et~al\mbox{.}}{2022b}]%
        {Chen2022CorpusBrainPA}
\bibfield{author}{\bibinfo{person}{Jiangui Chen}, \bibinfo{person}{Ruqing
  Zhang}, \bibinfo{person}{J. Guo}, \bibinfo{person}{Y. Liu},
  \bibinfo{person}{Yixing Fan}, {and} \bibinfo{person}{Xueqi Cheng}.}
  \bibinfo{year}{2022}\natexlab{b}.
\newblock \showarticletitle{CorpusBrain: Pre-train a Generative Retrieval Model
  for Knowledge-Intensive Language Tasks}. In
  \bibinfo{booktitle}{\emph{Proceedings of the 31st ACM International
  Conference on Information \& Knowledge Management}}.
\newblock


\bibitem[\protect\citeauthoryear{Chen, Liu, He, Sun, and Sun}{Chen
  et~al\mbox{.}}{2023a}]%
        {Chen2023UnderstandingDS}
\bibfield{author}{\bibinfo{person}{Xiaoyang Chen}, \bibinfo{person}{Yanjiang
  Liu}, \bibinfo{person}{Ben He}, \bibinfo{person}{Le Sun}, {and}
  \bibinfo{person}{Yingfei Sun}.} \bibinfo{year}{2023}\natexlab{a}.
\newblock \showarticletitle{Understanding Differential Search Index for Text
  Retrieval}.
\newblock \bibinfo{journal}{\emph{ArXiv}}  \bibinfo{volume}{abs/2305.02073}
  (\bibinfo{year}{2023}).
\newblock


\bibitem[\protect\citeauthoryear{Cho, Lei, Tan, and Bansal}{Cho
  et~al\mbox{.}}{2021}]%
        {DBLP:conf/icml/ChoLTB21}
\bibfield{author}{\bibinfo{person}{Jaemin Cho}, \bibinfo{person}{Jie Lei},
  \bibinfo{person}{Hao Tan}, {and} \bibinfo{person}{Mohit Bansal}.}
  \bibinfo{year}{2021}\natexlab{}.
\newblock \showarticletitle{Unifying Vision-and-Language Tasks via Text
  Generation}. In \bibinfo{booktitle}{\emph{Proceedings of the 38th
  International Conference on Machine Learning}}, Vol.~\bibinfo{volume}{139}.
\newblock


\bibitem[\protect\citeauthoryear{Christakopoulou, Radlinski, and
  Hofmann}{Christakopoulou et~al\mbox{.}}{2016}]%
        {Christakopoulou2016TowardsCR}
\bibfield{author}{\bibinfo{person}{Konstantina Christakopoulou},
  \bibinfo{person}{Filip Radlinski}, {and} \bibinfo{person}{Katja Hofmann}.}
  \bibinfo{year}{2016}\natexlab{}.
\newblock \showarticletitle{Towards Conversational Recommender Systems}. In
  \bibinfo{booktitle}{\emph{Proceedings of the 22nd ACM SIGKDD International
  Conference on Knowledge Discovery and Data Mining}}.
\newblock


\bibitem[\protect\citeauthoryear{Clarke, Craswell, and Soboroff}{Clarke
  et~al\mbox{.}}{2009}]%
        {Clarke2009OverviewOT}
\bibfield{author}{\bibinfo{person}{Charles L.~A. Clarke}, \bibinfo{person}{Nick
  Craswell}, {and} \bibinfo{person}{Ian Soboroff}.}
  \bibinfo{year}{2009}\natexlab{}.
\newblock \showarticletitle{Overview of the TREC 2009 Web Track}. In
  \bibinfo{booktitle}{\emph{Text REtrieval Conference}}.
\newblock


\bibitem[\protect\citeauthoryear{Clarke, Craswell, Soboroff, and
  Cormack}{Clarke et~al\mbox{.}}{2010}]%
        {Clarke2010OverviewOT}
\bibfield{author}{\bibinfo{person}{Charles L.~A. Clarke}, \bibinfo{person}{Nick
  Craswell}, \bibinfo{person}{Ian Soboroff}, {and} \bibinfo{person}{Gordon~V.
  Cormack}.} \bibinfo{year}{2010}\natexlab{}.
\newblock \showarticletitle{Overview of the TREC 2010 Web Track}. In
  \bibinfo{booktitle}{\emph{Text REtrieval Conference}}.
\newblock


\bibitem[\protect\citeauthoryear{Coden, Gruhl, Lewis, and Mendes}{Coden
  et~al\mbox{.}}{2015}]%
        {Coden2015DidYM}
\bibfield{author}{\bibinfo{person}{Anni Coden}, \bibinfo{person}{Daniel~F.
  Gruhl}, \bibinfo{person}{Neal Lewis}, {and} \bibinfo{person}{Pablo~N.
  Mendes}.} \bibinfo{year}{2015}\natexlab{}.
\newblock \showarticletitle{Did you mean A or B? Supporting Clarification
  Dialog for Entity Disambiguation}. In
  \bibinfo{booktitle}{\emph{SumPre-HSWI@ESWC}}.
\newblock


\bibitem[\protect\citeauthoryear{Datta, Varma, Chowdary, and Singh}{Datta
  et~al\mbox{.}}{2017}]%
        {Datta2017MultimodalRU}
\bibfield{author}{\bibinfo{person}{Deepanwita Datta}, \bibinfo{person}{Shubham
  Varma}, \bibinfo{person}{C.~Ravindranath Chowdary}, {and}
  \bibinfo{person}{Sanjay~Kumar Singh}.} \bibinfo{year}{2017}\natexlab{}.
\newblock \showarticletitle{Multimodal Retrieval using Mutual Information based
  Textual Query Reformulation}.
\newblock \bibinfo{journal}{\emph{Expert Systems with Applications}}
  \bibinfo{volume}{68} (\bibinfo{year}{2017}).
\newblock


\bibitem[\protect\citeauthoryear{Deldjoo, Trippas, and Zamani}{Deldjoo
  et~al\mbox{.}}{2021}]%
        {Deldjoo2021TowardsMC}
\bibfield{author}{\bibinfo{person}{Yashar Deldjoo}, \bibinfo{person}{Johanne~R.
  Trippas}, {and} \bibinfo{person}{Hamed Zamani}.}
  \bibinfo{year}{2021}\natexlab{}.
\newblock \showarticletitle{Towards Multi-Modal Conversational Information
  Seeking}. In \bibinfo{booktitle}{\emph{Proceedings of the 44th International
  ACM SIGIR Conference on Research and Development in Information Retrieval}}.
\newblock


\bibitem[\protect\citeauthoryear{Devlin, Chang, Lee, and Toutanova}{Devlin
  et~al\mbox{.}}{2019}]%
        {Devlin2019BERTPO}
\bibfield{author}{\bibinfo{person}{Jacob Devlin}, \bibinfo{person}{Ming-Wei
  Chang}, \bibinfo{person}{Kenton Lee}, {and} \bibinfo{person}{Kristina
  Toutanova}.} \bibinfo{year}{2019}\natexlab{}.
\newblock \showarticletitle{BERT: Pre-training of Deep Bidirectional
  Transformers for Language Understanding}.
\newblock \bibinfo{journal}{\emph{ArXiv}}  \bibinfo{volume}{abs/1810.04805}
  (\bibinfo{year}{2019}).
\newblock


\bibitem[\protect\citeauthoryear{Gao, Jin, Chen, Qiu, Wei, Hu, and Wang}{Gao
  et~al\mbox{.}}{2020}]%
        {Gao2020FashionBERTTA}
\bibfield{author}{\bibinfo{person}{Dehong Gao}, \bibinfo{person}{Linbo Jin},
  \bibinfo{person}{Ben Chen}, \bibinfo{person}{Minghui Qiu},
  \bibinfo{person}{Yi Wei}, \bibinfo{person}{Y. Hu}, {and}
  \bibinfo{person}{Haozhe~Jasper Wang}.} \bibinfo{year}{2020}\natexlab{}.
\newblock \showarticletitle{FashionBERT: Text and Image Matching with Adaptive
  Loss for Cross-modal Retrieval}. In \bibinfo{booktitle}{\emph{Proceedings of
  the 43rd International ACM SIGIR Conference on Research and Development in
  Information Retrieval}}.
\newblock


\bibitem[\protect\citeauthoryear{Hancock, Bordes, Mazar{\'e}, and
  Weston}{Hancock et~al\mbox{.}}{2019}]%
        {Hancock2019LearningFD}
\bibfield{author}{\bibinfo{person}{Braden Hancock}, \bibinfo{person}{Antoine
  Bordes}, \bibinfo{person}{Pierre-Emmanuel Mazar{\'e}}, {and}
  \bibinfo{person}{Jason Weston}.} \bibinfo{year}{2019}\natexlab{}.
\newblock \showarticletitle{Learning from Dialogue after Deployment: Feed
  Yourself, Chatbot!}. In \bibinfo{booktitle}{\emph{Proceedings of the 2019
  Annual Meeting of the Association for Computational Linguistics}}.
\newblock


\bibitem[\protect\citeauthoryear{Hashemi, Zamani, and Croft}{Hashemi
  et~al\mbox{.}}{2020}]%
        {Hashemi2020GuidedTL}
\bibfield{author}{\bibinfo{person}{Helia Hashemi}, \bibinfo{person}{Hamed
  Zamani}, {and} \bibinfo{person}{W.~Bruce Croft}.}
  \bibinfo{year}{2020}\natexlab{}.
\newblock \showarticletitle{Guided Transformer: Leveraging Multiple External
  Sources for Representation Learning in Conversational Search}. In
  \bibinfo{booktitle}{\emph{Proceedings of the 43rd International ACM SIGIR
  Conference on Research and Development in Information Retrieval}}.
\newblock


\bibitem[\protect\citeauthoryear{Hearst}{Hearst}{1999}]%
        {Hearst1999TrendsC}
\bibfield{author}{\bibinfo{person}{Marti~A. Hearst}.}
  \bibinfo{year}{1999}\natexlab{}.
\newblock \showarticletitle{Trends \& Controversies: Mixed-initiative
  Interaction}.
\newblock \bibinfo{journal}{\emph{IEEE Intell. Syst.}}  \bibinfo{volume}{14}
  (\bibinfo{year}{1999}).
\newblock


\bibitem[\protect\citeauthoryear{Kiesel, Bahrami, Stein, Anand, and
  Hagen}{Kiesel et~al\mbox{.}}{2018}]%
        {Kiesel2018TowardVQ}
\bibfield{author}{\bibinfo{person}{Johannes Kiesel}, \bibinfo{person}{Arefeh
  Bahrami}, \bibinfo{person}{Benno Stein}, \bibinfo{person}{Avishek Anand},
  {and} \bibinfo{person}{Matthias Hagen}.} \bibinfo{year}{2018}\natexlab{}.
\newblock \showarticletitle{Toward Voice Query Clarification}. In
  \bibinfo{booktitle}{\emph{Proceedings of the 41st International ACM SIGIR
  Conference on Research \& Development in Information Retrieval}}.
\newblock


\bibitem[\protect\citeauthoryear{Krishna, Zhu, Groth, Johnson, Hata, Kravitz,
  Chen, Kalantidis, Li, Shamma, Bernstein, and Fei-Fei}{Krishna
  et~al\mbox{.}}{2016}]%
        {Krishna2016VisualGC}
\bibfield{author}{\bibinfo{person}{Ranjay Krishna}, \bibinfo{person}{Yuke Zhu},
  \bibinfo{person}{Oliver Groth}, \bibinfo{person}{Justin Johnson},
  \bibinfo{person}{Kenji Hata}, \bibinfo{person}{Joshua Kravitz},
  \bibinfo{person}{Stephanie Chen}, \bibinfo{person}{Yannis Kalantidis},
  \bibinfo{person}{Li-Jia Li}, \bibinfo{person}{David~A. Shamma},
  \bibinfo{person}{Michael~S. Bernstein}, {and} \bibinfo{person}{Li Fei-Fei}.}
  \bibinfo{year}{2016}\natexlab{}.
\newblock \showarticletitle{Visual Genome: Connecting Language and Vision Using
  Crowdsourced Dense Image Annotations}.
\newblock \bibinfo{journal}{\emph{International Journal of Computer Vision}}
  \bibinfo{volume}{123} (\bibinfo{year}{2016}).
\newblock


\bibitem[\protect\citeauthoryear{Lee, Kim, Chang, Oh, Yang, Karpukhin, Lu, and
  Seo}{Lee et~al\mbox{.}}{2022}]%
        {Lee2022ContextualizedGR}
\bibfield{author}{\bibinfo{person}{Hyunji Lee}, \bibinfo{person}{Jaeyoung Kim},
  \bibinfo{person}{Hoyeon Chang}, \bibinfo{person}{Hanseok Oh},
  \bibinfo{person}{Sohee Yang}, \bibinfo{person}{Vladimir Karpukhin},
  \bibinfo{person}{Yi Lu}, {and} \bibinfo{person}{Minjoon Seo}.}
  \bibinfo{year}{2022}\natexlab{}.
\newblock \showarticletitle{Contextualized Generative Retrieval}.
\newblock \bibinfo{journal}{\emph{ArXiv}}  \bibinfo{volume}{abs/2210.02068}
  (\bibinfo{year}{2022}).
\newblock


\bibitem[\protect\citeauthoryear{Lester, Al-Rfou, and Constant}{Lester
  et~al\mbox{.}}{2021}]%
        {Lester2021ThePO}
\bibfield{author}{\bibinfo{person}{Brian Lester}, \bibinfo{person}{Rami
  Al-Rfou}, {and} \bibinfo{person}{Noah Constant}.}
  \bibinfo{year}{2021}\natexlab{}.
\newblock \showarticletitle{The Power of Scale for Parameter-Efficient Prompt
  Tuning}.
\newblock \bibinfo{journal}{\emph{ArXiv}}  \bibinfo{volume}{abs/2104.08691}
  (\bibinfo{year}{2021}).
\newblock


\bibitem[\protect\citeauthoryear{Li, Li, Savarese, and Hoi}{Li
  et~al\mbox{.}}{2023}]%
        {Li2023BLIP2BL}
\bibfield{author}{\bibinfo{person}{Junnan Li}, \bibinfo{person}{Dongxu Li},
  \bibinfo{person}{Silvio Savarese}, {and} \bibinfo{person}{Steven C.~H. Hoi}.}
  \bibinfo{year}{2023}\natexlab{}.
\newblock \showarticletitle{BLIP-2: Bootstrapping Language-Image Pre-training
  with Frozen Image Encoders and Large Language Models}.
\newblock \bibinfo{journal}{\emph{ArXiv}}  \bibinfo{volume}{abs/2301.12597}
  (\bibinfo{year}{2023}).
\newblock


\bibitem[\protect\citeauthoryear{Li, Yatskar, Yin, Hsieh, and Chang}{Li
  et~al\mbox{.}}{2019}]%
        {Li2019VisualBERTAS}
\bibfield{author}{\bibinfo{person}{Liunian~Harold Li}, \bibinfo{person}{Mark
  Yatskar}, \bibinfo{person}{Da Yin}, \bibinfo{person}{Cho-Jui Hsieh}, {and}
  \bibinfo{person}{Kai-Wei Chang}.} \bibinfo{year}{2019}\natexlab{}.
\newblock \showarticletitle{VisualBERT: A Simple and Performant Baseline for
  Vision and Language}.
\newblock \bibinfo{journal}{\emph{ArXiv}}  \bibinfo{volume}{abs/1908.03557}
  (\bibinfo{year}{2019}).
\newblock


\bibitem[\protect\citeauthoryear{Li, Yin, Li, Hu, Zhang, Zhang, Wang, Hu, Dong,
  Wei, Choi, and Gao}{Li et~al\mbox{.}}{2020}]%
        {Li2020OscarOA}
\bibfield{author}{\bibinfo{person}{Xiujun Li}, \bibinfo{person}{Xi Yin},
  \bibinfo{person}{Chunyuan Li}, \bibinfo{person}{Xiaowei Hu},
  \bibinfo{person}{Pengchuan Zhang}, \bibinfo{person}{Lei Zhang},
  \bibinfo{person}{Lijuan Wang}, \bibinfo{person}{Houdong Hu},
  \bibinfo{person}{Li Dong}, \bibinfo{person}{Furu Wei}, \bibinfo{person}{Yejin
  Choi}, {and} \bibinfo{person}{Jianfeng Gao}.}
  \bibinfo{year}{2020}\natexlab{}.
\newblock \showarticletitle{Oscar: Object-Semantics Aligned Pre-training for
  Vision-Language Tasks}. In \bibinfo{booktitle}{\emph{Proceedings of the 2020
  European Conference on Computer Vision}}.
\newblock


\bibitem[\protect\citeauthoryear{Liao, Long, Zhang, Huang, and Chua}{Liao
  et~al\mbox{.}}{2021}]%
        {Liao2021MMConvAE}
\bibfield{author}{\bibinfo{person}{Lizi Liao}, \bibinfo{person}{Le~Hong Long},
  \bibinfo{person}{Zheng Zhang}, \bibinfo{person}{Minlie Huang}, {and}
  \bibinfo{person}{Tat-Seng Chua}.} \bibinfo{year}{2021}\natexlab{}.
\newblock \showarticletitle{MMConv: An Environment for Multimodal
  Conversational Search across Multiple Domains}. In
  \bibinfo{booktitle}{\emph{Proceedings of the 44th International ACM SIGIR
  Conference on Research and Development in Information Retrieval}}.
\newblock


\bibitem[\protect\citeauthoryear{Liao, Ma, He, Hong, and Chua}{Liao
  et~al\mbox{.}}{2018}]%
        {Liao2018KnowledgeawareMD}
\bibfield{author}{\bibinfo{person}{Lizi Liao}, \bibinfo{person}{Yunshan Ma},
  \bibinfo{person}{Xiangnan He}, \bibinfo{person}{Richang Hong}, {and}
  \bibinfo{person}{Tat-Seng Chua}.} \bibinfo{year}{2018}\natexlab{}.
\newblock \showarticletitle{Knowledge-aware Multimodal Dialogue Systems}. In
  \bibinfo{booktitle}{\emph{Proceedings of the 26th ACM international
  conference on Multimedia}}.
\newblock


\bibitem[\protect\citeauthoryear{Ma, Kleemann, and Ziegler}{Ma
  et~al\mbox{.}}{2021}]%
        {Ma2021MixedModalityII}
\bibfield{author}{\bibinfo{person}{Yuan Ma}, \bibinfo{person}{Timm Kleemann},
  {and} \bibinfo{person}{J{\"u}rgen Ziegler}.} \bibinfo{year}{2021}\natexlab{}.
\newblock \showarticletitle{Mixed-Modality Interaction in Conversational
  Recommender Systems}. In \bibinfo{booktitle}{\emph{IntRS workshop at
  RecSys}}.
\newblock


\bibitem[\protect\citeauthoryear{MacAvaney, Yates, Cohan, and
  Goharian}{MacAvaney et~al\mbox{.}}{2019}]%
        {MacAvaney2019CEDRCE}
\bibfield{author}{\bibinfo{person}{Sean MacAvaney}, \bibinfo{person}{Andrew
  Yates}, \bibinfo{person}{Arman Cohan}, {and} \bibinfo{person}{Nazli
  Goharian}.} \bibinfo{year}{2019}\natexlab{}.
\newblock \showarticletitle{CEDR: Contextualized Embeddings for Document
  Ranking}. In \bibinfo{booktitle}{\emph{Proceedings of the 42nd International
  ACM SIGIR Conference on Research and Development in Information Retrieval}}.
\newblock


\bibitem[\protect\citeauthoryear{Meng, Ren, Chen, Ren, Xi, and de~Rijke}{Meng
  et~al\mbox{.}}{2021}]%
        {Meng2021InitiativeAwareSL}
\bibfield{author}{\bibinfo{person}{Chuan Meng}, \bibinfo{person}{Pengjie Ren},
  \bibinfo{person}{Zhumin Chen}, \bibinfo{person}{Zhaochun Ren},
  \bibinfo{person}{Tengxiao Xi}, {and} \bibinfo{person}{Maarten de Rijke}.}
  \bibinfo{year}{2021}\natexlab{}.
\newblock \showarticletitle{Initiative-Aware Self-Supervised Learning for
  Knowledge-Grounded Conversations}. In \bibinfo{booktitle}{\emph{Proceedings
  of the 44th International ACM SIGIR Conference on Research and Development in
  Information Retrieval}}.
\newblock


\bibitem[\protect\citeauthoryear{Mour{\~a}o, Martins, and
  Magalh{\~a}es}{Mour{\~a}o et~al\mbox{.}}{2015}]%
        {Mouro2015MultimodalMI}
\bibfield{author}{\bibinfo{person}{Andr{\'e} Mour{\~a}o},
  \bibinfo{person}{Fl{\'a}vio Martins}, {and} \bibinfo{person}{Jo{\~a}o
  Magalh{\~a}es}.} \bibinfo{year}{2015}\natexlab{}.
\newblock \showarticletitle{Multimodal Medical Information Retrieval with
  Unsupervised Rank Fusion}.
\newblock \bibinfo{journal}{\emph{Computerized Medical Imaging and Graphics:
  The Official Journal of the Computerized Medical Imaging Society}}
  \bibinfo{volume}{39} (\bibinfo{year}{2015}).
\newblock


\bibitem[\protect\citeauthoryear{Murrugarra-Llerena and
  Kovashka}{Murrugarra-Llerena and Kovashka}{2018}]%
        {MurrugarraLlerena2018ImageRW}
\bibfield{author}{\bibinfo{person}{Nils Murrugarra-Llerena} {and}
  \bibinfo{person}{Adriana Kovashka}.} \bibinfo{year}{2018}\natexlab{}.
\newblock \showarticletitle{Image Retrieval with Mixed Initiative and
  Multimodal Feedback}. In \bibinfo{booktitle}{\emph{British Machine Vision
  Conference}}.
\newblock


\bibitem[\protect\citeauthoryear{Narayan, Williams, Perugini, and
  Ramakrishnan}{Narayan et~al\mbox{.}}{2003}]%
        {Narayan2003StagingTF}
\bibfield{author}{\bibinfo{person}{Michael Narayan},
  \bibinfo{person}{Christopher Williams}, \bibinfo{person}{Saverio Perugini},
  {and} \bibinfo{person}{Naren Ramakrishnan}.} \bibinfo{year}{2003}\natexlab{}.
\newblock \showarticletitle{Staging transformations for multimodal web
  interaction management}.
\newblock \bibinfo{journal}{\emph{ArXiv}}  \bibinfo{volume}{cs.IR/0311029}
  (\bibinfo{year}{2003}).
\newblock


\bibitem[\protect\citeauthoryear{OpenAI}{OpenAI}{2023}]%
        {OpenAI2023GPT4TR}
\bibfield{author}{\bibinfo{person}{OpenAI}.} \bibinfo{year}{2023}\natexlab{}.
\newblock \showarticletitle{GPT-4 Technical Report}.
\newblock \bibinfo{journal}{\emph{ArXiv}}  \bibinfo{volume}{abs/2303.08774}
  (\bibinfo{year}{2023}).
\newblock


\bibitem[\protect\citeauthoryear{Owoicho, Sekulic, Aliannejadi, Dalton, and
  Crestani}{Owoicho et~al\mbox{.}}{2023}]%
        {Owoicho2023ExploitingSU}
\bibfield{author}{\bibinfo{person}{Paul Owoicho}, \bibinfo{person}{Ivan
  Sekulic}, \bibinfo{person}{Mohammad Aliannejadi},
  \bibinfo{person}{Jeffrey~Stephen Dalton}, {and} \bibinfo{person}{Fabio~A.
  Crestani}.} \bibinfo{year}{2023}\natexlab{}.
\newblock \showarticletitle{Exploiting Simulated User Feedback for
  Conversational Search: Ranking, Rewriting, and Beyond}.
\newblock \bibinfo{journal}{\emph{Proceedings of the 46th International ACM
  SIGIR Conference on Research and Development in Information Retrieval}}
  (\bibinfo{year}{2023}).
\newblock


\bibitem[\protect\citeauthoryear{Paszke, Gross, Massa, Lerer, Bradbury, Chanan,
  Killeen, Lin, Gimelshein, Antiga, Desmaison, K{\"o}pf, Yang, DeVito, Raison,
  Tejani, Chilamkurthy, Steiner, Fang, Bai, and Chintala}{Paszke
  et~al\mbox{.}}{2019}]%
        {Paszke2019PyTorchAI}
\bibfield{author}{\bibinfo{person}{Adam Paszke}, \bibinfo{person}{Sam Gross},
  \bibinfo{person}{Francisco Massa}, \bibinfo{person}{Adam Lerer},
  \bibinfo{person}{James Bradbury}, \bibinfo{person}{Gregory Chanan},
  \bibinfo{person}{Trevor Killeen}, \bibinfo{person}{Zeming Lin},
  \bibinfo{person}{Natalia Gimelshein}, \bibinfo{person}{Luca Antiga},
  \bibinfo{person}{Alban Desmaison}, \bibinfo{person}{Andreas K{\"o}pf},
  \bibinfo{person}{Edward Yang}, \bibinfo{person}{Zach DeVito},
  \bibinfo{person}{Martin Raison}, \bibinfo{person}{Alykhan Tejani},
  \bibinfo{person}{Sasank Chilamkurthy}, \bibinfo{person}{Benoit Steiner},
  \bibinfo{person}{Lu Fang}, \bibinfo{person}{Junjie Bai}, {and}
  \bibinfo{person}{Soumith Chintala}.} \bibinfo{year}{2019}\natexlab{}.
\newblock \showarticletitle{PyTorch: An Imperative Style, High-Performance Deep
  Learning Library}. In \bibinfo{booktitle}{\emph{Neural Information Processing
  Systems}}.
\newblock


\bibitem[\protect\citeauthoryear{Ponte and Croft}{Ponte and Croft}{1998}]%
        {Ponte1998ALM}
\bibfield{author}{\bibinfo{person}{Jay~M. Ponte} {and}
  \bibinfo{person}{W.~Bruce Croft}.} \bibinfo{year}{1998}\natexlab{}.
\newblock \showarticletitle{A Language Modeling Approach to Information
  Retrieval}.
\newblock \bibinfo{journal}{\emph{ACM SIGIR Forum}}  \bibinfo{volume}{51}
  (\bibinfo{year}{1998}).
\newblock


\bibitem[\protect\citeauthoryear{Qin and Liu}{Qin and Liu}{2013}]%
        {Qin2013IntroducingL4}
\bibfield{author}{\bibinfo{person}{Tao Qin} {and} \bibinfo{person}{Tie-Yan
  Liu}.} \bibinfo{year}{2013}\natexlab{}.
\newblock \showarticletitle{Introducing LETOR 4.0 Datasets}.
\newblock \bibinfo{journal}{\emph{ArXiv}}  \bibinfo{volume}{abs/1306.2597}
  (\bibinfo{year}{2013}).
\newblock


\bibitem[\protect\citeauthoryear{Quintano and Rodrigues}{Quintano and
  Rodrigues}{2008}]%
        {Quintano2008QuestionAnsweringCD}
\bibfield{author}{\bibinfo{person}{Luis Quintano} {and}
  \bibinfo{person}{Irene~Pimenta Rodrigues}.} \bibinfo{year}{2008}\natexlab{}.
\newblock \showarticletitle{Question/Answering Clarification Dialogues}. In
  \bibinfo{booktitle}{\emph{MICAI 2008: Advances in Artificial Intelligence}}.
\newblock


\bibitem[\protect\citeauthoryear{Radford, Kim, Hallacy, Ramesh, Goh, Agarwal,
  Sastry, Askell, Mishkin, Clark, Krueger, and Sutskever}{Radford
  et~al\mbox{.}}{2021}]%
        {Radford2021LearningTV}
\bibfield{author}{\bibinfo{person}{Alec Radford}, \bibinfo{person}{Jong~Wook
  Kim}, \bibinfo{person}{Chris Hallacy}, \bibinfo{person}{Aditya Ramesh},
  \bibinfo{person}{Gabriel Goh}, \bibinfo{person}{Sandhini Agarwal},
  \bibinfo{person}{Girish Sastry}, \bibinfo{person}{Amanda Askell},
  \bibinfo{person}{Pamela Mishkin}, \bibinfo{person}{Jack Clark},
  \bibinfo{person}{Gretchen Krueger}, {and} \bibinfo{person}{Ilya Sutskever}.}
  \bibinfo{year}{2021}\natexlab{}.
\newblock \showarticletitle{Learning Transferable Visual Models From Natural
  Language Supervision}. In \bibinfo{booktitle}{\emph{Proceedings of the 38th
  International Conference on Machine Learning}}.
\newblock


\bibitem[\protect\citeauthoryear{Raffel, Shazeer, Roberts, Lee, Narang, Matena,
  Zhou, Li, and Liu}{Raffel et~al\mbox{.}}{2019}]%
        {Raffel2019ExploringTL}
\bibfield{author}{\bibinfo{person}{Colin Raffel}, \bibinfo{person}{Noam~M.
  Shazeer}, \bibinfo{person}{Adam Roberts}, \bibinfo{person}{Katherine Lee},
  \bibinfo{person}{Sharan Narang}, \bibinfo{person}{Michael Matena},
  \bibinfo{person}{Yanqi Zhou}, \bibinfo{person}{Wei Li}, {and}
  \bibinfo{person}{Peter~J. Liu}.} \bibinfo{year}{2019}\natexlab{}.
\newblock \showarticletitle{Exploring the Limits of Transfer Learning with a
  Unified Text-to-Text Transformer}.
\newblock \bibinfo{journal}{\emph{ArXiv}}  \bibinfo{volume}{abs/1910.10683}
  (\bibinfo{year}{2019}).
\newblock


\bibitem[\protect\citeauthoryear{Rao and Daum{\'e}}{Rao and Daum{\'e}}{2018}]%
        {Rao2018LearningTA}
\bibfield{author}{\bibinfo{person}{Sudha Rao} {and} \bibinfo{person}{Hal
  Daum{\'e}}.} \bibinfo{year}{2018}\natexlab{}.
\newblock \showarticletitle{Learning to Ask Good Questions: Ranking
  Clarification Questions using Neural Expected Value of Perfect Information}.
  In \bibinfo{booktitle}{\emph{Proceedings of the 2018 Annual Meeting of the
  Association for Computational Linguistics}}.
\newblock


\bibitem[\protect\citeauthoryear{Ren, He, Girshick, and Sun}{Ren
  et~al\mbox{.}}{2015}]%
        {Ren2015FasterRT}
\bibfield{author}{\bibinfo{person}{Shaoqing Ren}, \bibinfo{person}{Kaiming He},
  \bibinfo{person}{Ross~B. Girshick}, {and} \bibinfo{person}{Jian Sun}.}
  \bibinfo{year}{2015}\natexlab{}.
\newblock \showarticletitle{Faster R-CNN: Towards Real-Time Object Detection
  with Region Proposal Networks}.
\newblock \bibinfo{journal}{\emph{IEEE Transactions on Pattern Analysis and
  Machine Intelligence}}  \bibinfo{volume}{39} (\bibinfo{year}{2015}).
\newblock


\bibitem[\protect\citeauthoryear{Robertson and Zaragoza}{Robertson and
  Zaragoza}{2009}]%
        {Robertson2009ThePR}
\bibfield{author}{\bibinfo{person}{Stephen~E. Robertson} {and}
  \bibinfo{person}{Hugo Zaragoza}.} \bibinfo{year}{2009}\natexlab{}.
\newblock \showarticletitle{The Probabilistic Relevance Framework: BM25 and
  Beyond}.
\newblock \bibinfo{journal}{\emph{Foundations and Trends in Information
  Retrieval}}  \bibinfo{volume}{3} (\bibinfo{year}{2009}).
\newblock


\bibitem[\protect\citeauthoryear{Saraiva, Cavalcanti, de~Moura, Gonçalves, and
  da~Silva~Torres}{Saraiva et~al\mbox{.}}{2016}]%
        {Saraiva2016AMQ}
\bibfield{author}{\bibinfo{person}{Patricia~Correia Saraiva},
  \bibinfo{person}{Jo{\~a}o M.~B. Cavalcanti}, \bibinfo{person}{Edleno~Silva de
  Moura}, \bibinfo{person}{Marcos~Andr{\'e} Gonçalves}, {and}
  \bibinfo{person}{Ricardo da Silva~Torres}.} \bibinfo{year}{2016}\natexlab{}.
\newblock \showarticletitle{A Multimodal Query Expansion Based on Genetic
  Programming for Visually-oriented E-commerce Applications}.
\newblock \bibinfo{journal}{\emph{Inf. Process. Manag.}}  \bibinfo{volume}{52}
  (\bibinfo{year}{2016}).
\newblock


\bibitem[\protect\citeauthoryear{Sekulic, Aliannejadi, and Crestani}{Sekulic
  et~al\mbox{.}}{2021}]%
        {Sekulic2021UserEP}
\bibfield{author}{\bibinfo{person}{Ivan Sekulic}, \bibinfo{person}{Mohammad
  Aliannejadi}, {and} \bibinfo{person}{Fabio~A. Crestani}.}
  \bibinfo{year}{2021}\natexlab{}.
\newblock \showarticletitle{User Engagement Prediction for Clarification in
  Search}. In \bibinfo{booktitle}{\emph{Proceedings of the 43rd European
  Conference on Information Retrieval}}.
\newblock


\bibitem[\protect\citeauthoryear{Srihari, Rao, Han, Srikanth, and Wu}{Srihari
  et~al\mbox{.}}{2000}]%
        {Srihari2000AMF}
\bibfield{author}{\bibinfo{person}{Rohini~K. Srihari}, \bibinfo{person}{Aibing
  Rao}, \bibinfo{person}{Benjamin Han}, \bibinfo{person}{Munirathnam Srikanth},
  {and} \bibinfo{person}{Xiaoyun Wu}.} \bibinfo{year}{2000}\natexlab{}.
\newblock \showarticletitle{A Model for Multimodal Information Retrieval}. In
  \bibinfo{booktitle}{\emph{Proceedings of the 2000 IEEE International
  Conference on Multimedia and Expo.}}, Vol.~\bibinfo{volume}{2}.
\newblock


\bibitem[\protect\citeauthoryear{Talmor, Yoran, Catav, Lahav, Wang, Asai,
  Ilharco, Hajishirzi, and Berant}{Talmor et~al\mbox{.}}{2021}]%
        {Talmor2021MultiModalQACQ}
\bibfield{author}{\bibinfo{person}{Alon Talmor}, \bibinfo{person}{Ori Yoran},
  \bibinfo{person}{Amnon Catav}, \bibinfo{person}{Dan Lahav},
  \bibinfo{person}{Yizhong Wang}, \bibinfo{person}{Akari Asai},
  \bibinfo{person}{Gabriel Ilharco}, \bibinfo{person}{Hannaneh Hajishirzi},
  {and} \bibinfo{person}{Jonathan Berant}.} \bibinfo{year}{2021}\natexlab{}.
\newblock \showarticletitle{MultiModalQA: Complex Question Answering over Text,
  Tables and Images}.
\newblock \bibinfo{journal}{\emph{ArXiv}}  \bibinfo{volume}{abs/2104.06039}
  (\bibinfo{year}{2021}).
\newblock


\bibitem[\protect\citeauthoryear{Tautkute and Trzciński}{Tautkute and
  Trzciński}{2021}]%
        {Tautkute2021IWT}
\bibfield{author}{\bibinfo{person}{Ivona Tautkute} {and}
  \bibinfo{person}{Tomasz Trzciński}.} \bibinfo{year}{2021}\natexlab{}.
\newblock \showarticletitle{I Want This Product but Different : Multimodal
  Retrieval with Synthetic Query Expansion}.
\newblock \bibinfo{journal}{\emph{ArXiv}}  \bibinfo{volume}{abs/2102.08871}
  (\bibinfo{year}{2021}).
\newblock


\bibitem[\protect\citeauthoryear{Tavakoli, Trippas, Zamani, Scholer, and
  Sanderson}{Tavakoli et~al\mbox{.}}{2022}]%
        {Tavakoli2022MIMICSDuoO}
\bibfield{author}{\bibinfo{person}{Leila Tavakoli}, \bibinfo{person}{Johanne~R.
  Trippas}, \bibinfo{person}{Hamed Zamani}, \bibinfo{person}{Falk Scholer},
  {and} \bibinfo{person}{Mark Sanderson}.} \bibinfo{year}{2022}\natexlab{}.
\newblock \showarticletitle{MIMICS-Duo: Offline \& Online Evaluation of Search
  Clarification}. In \bibinfo{booktitle}{\emph{Proceedings of the 45th
  International ACM SIGIR Conference on Research and Development in Information
  Retrieval}}.
\newblock


\bibitem[\protect\citeauthoryear{Tay, Tran, Dehghani, Ni, Bahri, Mehta, Qin,
  Hui, Zhao, Gupta, Schuster, Cohen, and Metzler}{Tay et~al\mbox{.}}{2022}]%
        {Tay2022TransformerMA}
\bibfield{author}{\bibinfo{person}{Yi Tay}, \bibinfo{person}{Vinh~Quang Tran},
  \bibinfo{person}{Mostafa Dehghani}, \bibinfo{person}{Jianmo Ni},
  \bibinfo{person}{Dara Bahri}, \bibinfo{person}{Harsh Mehta},
  \bibinfo{person}{Zhen Qin}, \bibinfo{person}{Kai Hui}, \bibinfo{person}{Zhe
  Zhao}, \bibinfo{person}{Jai Gupta}, \bibinfo{person}{Tal Schuster},
  \bibinfo{person}{William~W. Cohen}, {and} \bibinfo{person}{Donald Metzler}.}
  \bibinfo{year}{2022}\natexlab{}.
\newblock \showarticletitle{Transformer Memory as a Differentiable Search
  Index}.
\newblock \bibinfo{journal}{\emph{ArXiv}}  \bibinfo{volume}{abs/2202.06991}
  (\bibinfo{year}{2022}).
\newblock


\bibitem[\protect\citeauthoryear{Touvron, Lavril, Izacard, Martinet, Lachaux,
  Lacroix, Rozi{\`e}re, Goyal, Hambro, Azhar, Rodriguez, Joulin, Grave, and
  Lample}{Touvron et~al\mbox{.}}{2023}]%
        {Touvron2023LLaMAOA}
\bibfield{author}{\bibinfo{person}{Hugo Touvron}, \bibinfo{person}{Thibaut
  Lavril}, \bibinfo{person}{Gautier Izacard}, \bibinfo{person}{Xavier
  Martinet}, \bibinfo{person}{Marie-Anne Lachaux},
  \bibinfo{person}{Timoth{\'e}e Lacroix}, \bibinfo{person}{Baptiste
  Rozi{\`e}re}, \bibinfo{person}{Naman Goyal}, \bibinfo{person}{Eric Hambro},
  \bibinfo{person}{Faisal Azhar}, \bibinfo{person}{Aurelien Rodriguez},
  \bibinfo{person}{Armand Joulin}, \bibinfo{person}{Edouard Grave}, {and}
  \bibinfo{person}{Guillaume Lample}.} \bibinfo{year}{2023}\natexlab{}.
\newblock \showarticletitle{LLaMA: Open and Efficient Foundation Language
  Models}.
\newblock \bibinfo{journal}{\emph{ArXiv}}  \bibinfo{volume}{abs/2302.13971}
  (\bibinfo{year}{2023}).
\newblock


\bibitem[\protect\citeauthoryear{Tsagkias, King, Kallumadi, Murdock, and
  de~Rijke}{Tsagkias et~al\mbox{.}}{2020}]%
        {tsagkias-2020-challenges}
\bibfield{author}{\bibinfo{person}{Manos Tsagkias},
  \bibinfo{person}{Tracy~Holloway King}, \bibinfo{person}{Surya Kallumadi},
  \bibinfo{person}{Vanessa Murdock}, {and} \bibinfo{person}{Maarten de Rijke}.}
  \bibinfo{year}{2020}\natexlab{}.
\newblock \showarticletitle{Challenges and Research Opportunities in eCommerce
  Search and Recommendations}.
\newblock \bibinfo{journal}{\emph{SIGIR Forum}}  \bibinfo{volume}{54}
  (\bibinfo{year}{2020}).
\newblock


\bibitem[\protect\citeauthoryear{Vakulenko, Kanoulas, and de~Rijke}{Vakulenko
  et~al\mbox{.}}{2020}]%
        {Vakulenko2020AnAO}
\bibfield{author}{\bibinfo{person}{Svitlana Vakulenko},
  \bibinfo{person}{Evangelos Kanoulas}, {and} \bibinfo{person}{Maarten de
  Rijke}.} \bibinfo{year}{2020}\natexlab{}.
\newblock \showarticletitle{An Analysis of Mixed Initiative and Collaboration
  in Information-Seeking Dialogues}. In \bibinfo{booktitle}{\emph{Proceedings
  of the 43rd International ACM SIGIR Conference on Research and Development in
  Information Retrieval}}.
\newblock


\bibitem[\protect\citeauthoryear{Wang, Lin, Feng, He, and Chua}{Wang
  et~al\mbox{.}}{2023}]%
        {Wang2023GenerativeRT}
\bibfield{author}{\bibinfo{person}{Wenjie Wang}, \bibinfo{person}{Xinyu Lin},
  \bibinfo{person}{Fuli Feng}, \bibinfo{person}{Xiangnan He}, {and}
  \bibinfo{person}{Tat-Seng Chua}.} \bibinfo{year}{2023}\natexlab{}.
\newblock \showarticletitle{Generative Recommendation: Towards Next-generation
  Recommender Paradigm}.
\newblock \bibinfo{journal}{\emph{ArXiv}}  \bibinfo{volume}{abs/2304.03516}
  (\bibinfo{year}{2023}).
\newblock


\bibitem[\protect\citeauthoryear{Wang, Hou, Wang, Miao, Wu, Sun, Chen, Xia,
  Chi, Zhao, Liu, Xie, Sun, Deng, Zhang, and Yang}{Wang et~al\mbox{.}}{2022}]%
        {Wang2022ANC}
\bibfield{author}{\bibinfo{person}{Yujing Wang}, \bibinfo{person}{Ying Hou},
  \bibinfo{person}{Hong Wang}, \bibinfo{person}{Ziming Miao},
  \bibinfo{person}{Shibin Wu}, \bibinfo{person}{Hao Sun}, \bibinfo{person}{Qi
  Chen}, \bibinfo{person}{Yuqing Xia}, \bibinfo{person}{Chengmin Chi},
  \bibinfo{person}{Guoshuai Zhao}, \bibinfo{person}{Zheng Liu},
  \bibinfo{person}{Xing Xie}, \bibinfo{person}{Hao Sun},
  \bibinfo{person}{Weiwei Deng}, \bibinfo{person}{Qi Zhang}, {and}
  \bibinfo{person}{Mao Yang}.} \bibinfo{year}{2022}\natexlab{}.
\newblock \showarticletitle{A Neural Corpus Indexer for Document Retrieval}.
\newblock \bibinfo{journal}{\emph{ArXiv}}  \bibinfo{volume}{abs/2206.02743}
  (\bibinfo{year}{2022}).
\newblock


\bibitem[\protect\citeauthoryear{Wolf, Debut, Sanh, Chaumond, Delangue, Moi,
  Cistac, Rault, Louf, Funtowicz, and Brew}{Wolf et~al\mbox{.}}{2019}]%
        {Wolf2019HuggingFacesTS}
\bibfield{author}{\bibinfo{person}{Thomas Wolf}, \bibinfo{person}{Lysandre
  Debut}, \bibinfo{person}{Victor Sanh}, \bibinfo{person}{Julien Chaumond},
  \bibinfo{person}{Clement Delangue}, \bibinfo{person}{Anthony Moi},
  \bibinfo{person}{Pierric Cistac}, \bibinfo{person}{Tim Rault},
  \bibinfo{person}{R{\'e}mi Louf}, \bibinfo{person}{Morgan Funtowicz}, {and}
  \bibinfo{person}{Jamie Brew}.} \bibinfo{year}{2019}\natexlab{}.
\newblock \showarticletitle{HuggingFace's Transformers: State-of-the-art
  Natural Language Processing}.
\newblock \bibinfo{journal}{\emph{ArXiv}}  \bibinfo{volume}{abs/1910.03771}
  (\bibinfo{year}{2019}).
\newblock


\bibitem[\protect\citeauthoryear{Wu, Burges, Svore, and Gao}{Wu
  et~al\mbox{.}}{2010}]%
        {Wu2010AdaptingBF}
\bibfield{author}{\bibinfo{person}{Qiang Wu}, \bibinfo{person}{Christopher
  J.~C. Burges}, \bibinfo{person}{Krysta~Marie Svore}, {and}
  \bibinfo{person}{Jianfeng Gao}.} \bibinfo{year}{2010}\natexlab{}.
\newblock \showarticletitle{Adapting boosting for information retrieval
  measures}.
\newblock \bibinfo{journal}{\emph{Information Retrieval}}  \bibinfo{volume}{13}
  (\bibinfo{year}{2010}).
\newblock


\bibitem[\protect\citeauthoryear{Xie, Mao, de~Rijke, Zhang, Zhang, and Ma}{Xie
  et~al\mbox{.}}{2018}]%
        {Xie2018ConstructingAI}
\bibfield{author}{\bibinfo{person}{Xiaohui Xie}, \bibinfo{person}{Jiaxin Mao},
  \bibinfo{person}{M. de Rijke}, \bibinfo{person}{Ruizhe Zhang},
  \bibinfo{person}{Min Zhang}, {and} \bibinfo{person}{Shaoping Ma}.}
  \bibinfo{year}{2018}\natexlab{}.
\newblock \showarticletitle{Constructing an Interaction Behavior Model for Web
  Image Search}. In \bibinfo{booktitle}{\emph{Proceedings of the 41st
  International ACM SIGIR Conference on Research \& Development in Information
  Retrieval}}.
\newblock


\bibitem[\protect\citeauthoryear{Xu, Wang, Tang, Duan, Yang, Zeng, Zhou, and
  Sun}{Xu et~al\mbox{.}}{2019}]%
        {Xu2019AskingCQ}
\bibfield{author}{\bibinfo{person}{Jingjing Xu}, \bibinfo{person}{Yuechen
  Wang}, \bibinfo{person}{Duyu Tang}, \bibinfo{person}{Nan Duan},
  \bibinfo{person}{Pengcheng Yang}, \bibinfo{person}{Qi Zeng},
  \bibinfo{person}{Ming Zhou}, {and} \bibinfo{person}{Xu Sun}.}
  \bibinfo{year}{2019}\natexlab{}.
\newblock \showarticletitle{Asking Clarification Questions in Knowledge-Based
  Question Answering}. In \bibinfo{booktitle}{\emph{Proceedings of the 2019
  Conference on Empirical Methods in Natural Language Processing}}.
\newblock


\bibitem[\protect\citeauthoryear{Yin, Li, Lu, and Zhang}{Yin
  et~al\mbox{.}}{2019}]%
        {Yin2019EnhancingFR}
\bibfield{author}{\bibinfo{person}{Ruiping Yin}, \bibinfo{person}{Kan Li},
  \bibinfo{person}{Jie Lu}, {and} \bibinfo{person}{Guangquan Zhang}.}
  \bibinfo{year}{2019}\natexlab{}.
\newblock \showarticletitle{Enhancing Fashion Recommendation with Visual
  Compatibility Relationship}. In \bibinfo{booktitle}{\emph{Proceedings of the
  2019 World Wide Web Conference}}.
\newblock


\bibitem[\protect\citeauthoryear{Yuan and Lam}{Yuan and Lam}{2021}]%
        {Yuan2021ConversationalFI}
\bibfield{author}{\bibinfo{person}{Yifei Yuan} {and} \bibinfo{person}{Wai
  Lam}.} \bibinfo{year}{2021}\natexlab{}.
\newblock \showarticletitle{Conversational Fashion Image Retrieval via
  Multiturn Natural Language Feedback}.
\newblock \bibinfo{journal}{\emph{Proceedings of the 44th International ACM
  SIGIR Conference on Research and Development in Information Retrieval}}
  (\bibinfo{year}{2021}).
\newblock


\bibitem[\protect\citeauthoryear{Yuan, Shi, Wang, Chen, Jiang, You, and
  Lam}{Yuan et~al\mbox{.}}{2022}]%
        {Yuan2022McQueenAB}
\bibfield{author}{\bibinfo{person}{Yifei Yuan}, \bibinfo{person}{Chen Shi},
  \bibinfo{person}{Runze Wang}, \bibinfo{person}{Liyi Chen},
  \bibinfo{person}{Feijun Jiang}, \bibinfo{person}{Yuan You}, {and}
  \bibinfo{person}{Wai Lam}.} \bibinfo{year}{2022}\natexlab{}.
\newblock \showarticletitle{McQueen: A Benchmark for Multimodal Conversational
  Query Rewrite}. In \bibinfo{booktitle}{\emph{Proceedings of the 2022
  Conference on Empirical Methods in Natural Language Processing}}.
\newblock


\bibitem[\protect\citeauthoryear{Zamani, Dumais, Craswell, Bennett, and
  Lueck}{Zamani et~al\mbox{.}}{2020a}]%
        {Zamani2020GeneratingCQ}
\bibfield{author}{\bibinfo{person}{Hamed Zamani}, \bibinfo{person}{Susan~T.
  Dumais}, \bibinfo{person}{Nick Craswell}, \bibinfo{person}{Paul~N. Bennett},
  {and} \bibinfo{person}{Gord Lueck}.} \bibinfo{year}{2020}\natexlab{a}.
\newblock \showarticletitle{Generating Clarifying Questions for Information
  Retrieval}. In \bibinfo{booktitle}{\emph{Proceedings of The Web Conference
  2020}}.
\newblock


\bibitem[\protect\citeauthoryear{Zamani, Lueck, Chen, Quispe, Luu, and
  Craswell}{Zamani et~al\mbox{.}}{2020b}]%
        {Zamani2020MIMICSAL}
\bibfield{author}{\bibinfo{person}{Hamed Zamani}, \bibinfo{person}{Gord Lueck},
  \bibinfo{person}{Everest Chen}, \bibinfo{person}{Rodolfo Quispe},
  \bibinfo{person}{Flint Luu}, {and} \bibinfo{person}{Nick Craswell}.}
  \bibinfo{year}{2020}\natexlab{b}.
\newblock \showarticletitle{MIMICS: A Large-Scale Data Collection for Search
  Clarification}. In \bibinfo{booktitle}{\emph{Proceedings of the 29th ACM
  International Conference on Information \& Knowledge Management}}.
\newblock


\bibitem[\protect\citeauthoryear{Zamani, Mitra, Chen, Lueck, Diaz, Bennett,
  Craswell, and Dumais}{Zamani et~al\mbox{.}}{2020c}]%
        {Zamani2020AnalyzingAL}
\bibfield{author}{\bibinfo{person}{Hamed Zamani}, \bibinfo{person}{Bhaskar
  Mitra}, \bibinfo{person}{Everest Chen}, \bibinfo{person}{Gord Lueck},
  \bibinfo{person}{Fernando Diaz}, \bibinfo{person}{Paul~N. Bennett},
  \bibinfo{person}{Nick Craswell}, {and} \bibinfo{person}{Susan~T. Dumais}.}
  \bibinfo{year}{2020}\natexlab{c}.
\newblock \showarticletitle{Analyzing and Learning from User Interactions for
  Search Clarification}. In \bibinfo{booktitle}{\emph{Proceedings of the 43rd
  International ACM SIGIR Conference on Research and Development in Information
  Retrieval}}.
\newblock


\bibitem[\protect\citeauthoryear{Zhang, Li, Hu, Yang, Zhang, Wang, Choi, and
  Gao}{Zhang et~al\mbox{.}}{2021}]%
        {Zhang2021VinVLRV}
\bibfield{author}{\bibinfo{person}{Pengchuan Zhang}, \bibinfo{person}{Xiujun
  Li}, \bibinfo{person}{Xiaowei Hu}, \bibinfo{person}{Jianwei Yang},
  \bibinfo{person}{Lei Zhang}, \bibinfo{person}{Lijuan Wang},
  \bibinfo{person}{Yejin Choi}, {and} \bibinfo{person}{Jianfeng Gao}.}
  \bibinfo{year}{2021}\natexlab{}.
\newblock \showarticletitle{VinVL: Revisiting Visual Representations in
  Vision-Language Models}. In \bibinfo{booktitle}{\emph{Proceedings of the 2021
  IEEE/CVF Conference on Computer Vision and Pattern Recognition}}.
\newblock


\bibitem[\protect\citeauthoryear{Zhang, Chen, Ai, Yang, and Croft}{Zhang
  et~al\mbox{.}}{2018}]%
        {Zhang2018TowardsCS}
\bibfield{author}{\bibinfo{person}{Yongfeng Zhang}, \bibinfo{person}{Xu Chen},
  \bibinfo{person}{Qingyao Ai}, \bibinfo{person}{Liu Yang}, {and}
  \bibinfo{person}{W.~Bruce Croft}.} \bibinfo{year}{2018}\natexlab{}.
\newblock \showarticletitle{Towards Conversational Search and Recommendation:
  System Ask, User Respond}. In \bibinfo{booktitle}{\emph{Proceedings of the
  27th ACM International Conference on Information and Knowledge Management}}.
\newblock


\bibitem[\protect\citeauthoryear{Zhou, Palangi, Zhang, Hu, Corso, and Gao}{Zhou
  et~al\mbox{.}}{2019}]%
        {Zhou2019UnifiedVP}
\bibfield{author}{\bibinfo{person}{Luowei Zhou}, \bibinfo{person}{Hamid
  Palangi}, \bibinfo{person}{Lei Zhang}, \bibinfo{person}{Houdong Hu},
  \bibinfo{person}{Jason~J. Corso}, {and} \bibinfo{person}{Jianfeng Gao}.}
  \bibinfo{year}{2019}\natexlab{}.
\newblock \showarticletitle{Unified Vision-Language Pre-Training for Image
  Captioning and VQA}.
\newblock \bibinfo{journal}{\emph{ArXiv}}  \bibinfo{volume}{abs/1909.11059}
  (\bibinfo{year}{2019}).
\newblock


\bibitem[\protect\citeauthoryear{Zhou, Yao, Dou, Wu, and rong Wen}{Zhou
  et~al\mbox{.}}{2022}]%
        {Zhou2022DynamicRetrieverAP}
\bibfield{author}{\bibinfo{person}{Yujia Zhou}, \bibinfo{person}{Jing Yao},
  \bibinfo{person}{Zhicheng Dou}, \bibinfo{person}{Ledell~Yu Wu}, {and}
  \bibinfo{person}{Ji rong Wen}.} \bibinfo{year}{2022}\natexlab{}.
\newblock \showarticletitle{DynamicRetriever: A Pre-training Model-based IR
  System with Neither Sparse nor Dense Index}.
\newblock \bibinfo{journal}{\emph{ArXiv}}  \bibinfo{volume}{abs/2203.00537}
  (\bibinfo{year}{2022}).
\newblock


\bibitem[\protect\citeauthoryear{Zou, Aliannejadi, Kanoulas, Pera, and Liu}{Zou
  et~al\mbox{.}}{2022}]%
        {Zou2022UsersMC}
\bibfield{author}{\bibinfo{person}{Jie Zou}, \bibinfo{person}{Mohammad
  Aliannejadi}, \bibinfo{person}{E. Kanoulas}, \bibinfo{person}{Maria~Soledad
  Pera}, {and} \bibinfo{person}{Yiqun Liu}.} \bibinfo{year}{2022}\natexlab{}.
\newblock \showarticletitle{Users Meet Clarifying Questions: Toward a Better
  Understanding of User Interactions for Search Clarification}.
\newblock \bibinfo{journal}{\emph{ACM Transactions on Information Systems}}
  \bibinfo{volume}{41} (\bibinfo{year}{2022}).
\newblock


\end{thebibliography}
\appendix
\section{\OurData Quality control}
\label{qualitycontrol}
We implement several quality control measures. Firstly, we utilize the quality control mechanisms provided by Appen and \ac{AMT}. We require \ac{AMT} workers to have at least 10,000 approved \acp{HIT} and a lifetime approval rate greater than 97$\%$. We administer an onboarding test to all annotators to ensure their understanding of the task.
After \textbf{Phase 1} and \textbf{Phase 2}, two quality checkers are employed and instructed to evaluate the quality of the returned questions and images by considering three criteria: 
\begin{enumerate*}[label=(\roman*)]
    \item relevance of the clarifying question to the topic, 
    \item suitability of the clarifying question to be accompanied by images, and
    \item relevance of the images to the clarifying question.  
\end{enumerate*}
We require that the clarifying questions satisfy all three criteria to be included in the dataset; 2.1\% of the questions are marked as erroneous or invalid and are removed from our dataset. For answer submissions, we conduct manual checks by randomly  examining 10\% of the submissions. 

To ensure the overall quality of the data collection stages, a subset of 200 clarifying questions, along with their corresponding information need, facet, and images, are sampled. These samples are then independently assessed by three crowdworkers. The evaluation focuses on determining the suitability of the questions for image attachment and the relevance of the attached images to the user's information need. The results indicate that 98\% of the questions were judged as suitable for image attachment, while 96\% of the images were found to be relevant to the user's information need.
The small margin of difference between the suitability of questions for image attachment and the relevance of the attached images to the information need can be attributed to the explicit instructions given to the annotators to provide diverse images. While this might have led to some images being topically relevant but not directly aligned with the user's specific information need, the small margin indicates overall, the dataset collected is of good quality.

\section{\OurData Question term comparison} 
\label{termcompare}
\begin{table}[]
    \centering
    \caption{Top-8 key phrases in set 1 \& 2 questions.}
    \begin{tabular}{cp{6.2cm}}
    \toprule
    Set 1  & know, looking for, want, interested in, specific, information, referring to, see\\
    Set 2  & want, see, interested in, looking for, know, information, see photos, different \\
    \bottomrule
    \end{tabular}
    \label{tab:keyphrase}
\end{table}
To examine potential biases in the collection of multimodal questions, we conduct a comparative analysis of the language usage in ClariQ questions~(\textbf{set 1 questions}) and \ac{AMT} questions~(\textbf{set 2 questions}).  Table \ref{tab:keyphrase} presents the top-8 key phrases extracted from both sets. There is a notable overlap in the distribution of key phrases between the two sets with some words (e.g., ``want,'' ``know'') appearing frequently in both cases. Both sets of questions exhibit references to vision-related aspects, as indicated by the presence of terms like ``see.'' Set 2 questions demonstrate a higher inclination towards vision-related phrases, such as ``would you like to see'' or ``do you want photos,'' suggesting a stronger focus on image-related inquiries.

\section{Experimental setup}
\label{setup}
\header{Dataset.} 
Following~\cite{Aliannejadi2020ConvAI3GC}, we split our dataset into training/vali\-dation/test sets on different facets with 80\%/10\%/10\% proportions, resulting in 856/107/107 facets in each set. The training, validation, and test set contain 14,187/1,851/1,865 samples, respectively.

\header{Evaluation metrics and statistical test} 
For document retrieval, following~\cite{AliannejadiSigir19,Clarke2010OverviewOT}, we adopt \ac{MRR}, precision (P@k), normalized discounted cumulative gain (nDCG@k), expected reciprocal rank (ERR@k) where $k\in\{1,3,5\}$ as evaluation metrics. The ground-truth relevance documents are obtained from TREC and adjusted on the facet level following~\cite{AliannejadiSigir19}. We perform statistical significance testing using a two-tailed paired t-test with Bonferroni correction at a 99.9\% confidence interval ($p<0.001$).

\header{Compared methods}
To demonstrate the effectiveness of multimodal content in query clarification, we adopt various competitive baselines. We first consider several lexical methods:
\begin{itemize}[leftmargin=*,nosep]
\item \textbf{OriginalQuery}~\cite{AliannejadiSigir19} reports the performance where  retrieval is performed only on the user query without clarification.

\item \textbf{QL}~\cite{Ponte1998ALM,AliannejadiSigir19} is a \ac{QL} retrieval model that assigns different weights to the query, clarifying question and answer.

\item \textbf{BM25}~\cite{Robertson2009ThePR} is used to directly retrieve and rank the documents given the query, clarifying question, and answer.

\item \textbf{LambdaMART}~\cite{Wu2010AdaptingBF} is a \ac{LTR} baseline that learns to rank the documents according to queries. We use the 46 features listed in~\cite{Qin2013IntroducingL4}.
\end{itemize}

\noindent We then adopt the following  pipeline-based methods:
\begin{itemize}[leftmargin=*,nosep]
\item \textbf{BERT+CLIP+CLIP} adopts  the BERT model to perform question classification and the CLIP model for the image selection. For document retrieval, we also utilize CLIP retrieval for encoding all images and questions, as well as all document identifiers.
\item \textbf{BERT+CLIP+VLT5} utilizes the BERT model to perform question classification, CLIP model for image selection and the VLT5 model for document retrieval. 
\end{itemize}

Next, we adopt baselines also under a multi-task framework with the first two in unimodal and the rest in multimodal scenario. For unimodal baselines, we adopt BERT and T5 respectively as the base model to retrieve the documents according to the topic and text-only clarifying question:
\begin{itemize}[leftmargin=*,nosep]
\item \textbf{BERT}~\cite{Devlin2019BERTPO} shows the performance under the unimodal BERT ranking model, where clarifying questions are text-only. We use the BERT ranking implementation released in~\cite{MacAvaney2019CEDRCE}. 

\item \textbf{T5}~\cite{Raffel2019ExploringTL} adopts a generative retrieval setting to retrieve the documents by the text-only questions. The  generative model is trained to generate the keyword sequence of the relevant documents and can be seen as the text-only version of \OurModel.


\item \textbf{VisualBERT} is the MQC pipeline with the same training tasks as \OurModel but based on VisualBERT model.

\item \textbf{VisualBERT\textunderscore{w/o QC}}~\cite{Li2019VisualBERTAS} takes the query, the clarifying question with images, and the answer as input, without performing multimodal-enhanced question classification. 
\end{itemize}

\noindent Finally, we compare \OurModel with multiple variants to evaluate the efficacy of our model's design.

\begin{itemize}
[leftmargin=*,nosep]
\item \textbf{\OurModel} is our model described in Section \ref{model}.

\item \textbf{\OurModelNS\textunderscore{w/o QC}} is a variant of our model where the multimodal-enhanced question classification module is removed. 
\item 
\textbf{\OurModelNS\textunderscore{random-image}} is another  \OurModel variant where we randomly attach an image from image pool in the retrieval process.
\item \textbf{\OurModelNS\textunderscore{trel-image}} is a \OurModel variant where we attach the topic-relevant images rather than question-relevant images.
\item \textbf{\OurModelNS\textunderscore{oracle-best-image}} is an oracle variant of our model where the model always selects the best image.
\end{itemize}

\begin{table}[b]
    \caption{Performance of the image selection module. }
    \label{appendixism}
    \centering
    \small
    \setlength{\tabcolsep}{1.3mm}
    \begin{tabular}{cccc}
    \toprule
    \textbf{Method} & \textbf{Precision} &	\textbf{Relevancy} &	\textbf{Accuracy}\\
    \midrule
   Image Selection & 89.34 &	91.30	& 80.37 \\
   \bottomrule
    \end{tabular}
    \vspace{-1mm}
\end{table}

\begin{table}[b]
    \caption{Performance of the multimodal-enhanced question classification module.}
    \label{appendixmqc}
    \centering
    \small
    \setlength{\tabcolsep}{1.3mm}
    \begin{tabular}{cccc}
    \toprule
    \textbf{Method} & \textbf{P} &	\textbf{R} &	\textbf{F1}\\
    \midrule
    Bert& 	\textbf{90.65} &	85.81 &	\textbf{88.16} \\
    T5 (Ours)&	88.58	& \textbf{86.68}	 & 87.62 \\
   \bottomrule
    \end{tabular}
    \vspace{-1mm}
\end{table}

\header{Hyperparameter settings} 
Our code is based on Pytorch~\cite{Paszke2019PyTorchAI} and Huggingface Transformers~\cite{Wolf2019HuggingFacesTS}. For VLT5, we use the pre-trained base version. By default, we set the batch size to 32 and the learning rate to 5e-5, the model is fine-tuned for 5 runs with 30 epochs per run, each with different random seeds. In the test stage, all models decode words with beam size 15. For first-stage document retrieval, we use BM25~\cite{Robertson2009ThePR} to retrieve 100 documents for each facet. We only report the result on the second-stage re-ranking.

\begin{table*}[t]
    \caption{\OurModel performance with different document identifier strategies. }
    \label{appendixdocid}
    \centering
    \small
    \setlength{\tabcolsep}{1.5mm}
    \begin{tabular}{ccccccccccc}
    \toprule
Strategy &	MRR &	P@1	& P@3	& P@5	& nDCG@1	& nDCG@3	& nDCG@5 &	ERR@1	& ERR@3& 	ERR@5 \\
\midrule

$Doc_N$ &	50.23 &	48.90 &	36.58 &	33.45 &	27.18 &	20.83 &	18.17 &	12.66 &	16.07 &	18.25 \\
$Doc_{F5}$ &	53.47 &	51.95 &	38.22 &	36.01	&28.82	&23.60&	23.22	&14.34	&18.70&	20.71 \\
$Doc_K$ (Ours)&	\textbf{54.70}	&\textbf{53.38}&	\textbf{40.47}&	\textbf{36.65}&	\textbf{30.66}&	\textbf{24.57}&	\textbf{23.81}	&\textbf{15.63}&	\textbf{20.21}&	\textbf{21.64} \\
   \bottomrule
    \end{tabular}
    \vspace{-1mm}
\end{table*}

\begin{table*}[t]
    \caption{\OurModel performance performance with different backbones. }
    \label{appendixbackbone}
    \centering
    \small 
    \setlength{\tabcolsep}{1.5mm}
    \begin{tabular}{ccccccccccc}
    \toprule
Strategy &	MRR &	P@1	& P@3	& P@5	& nDCG@1	& nDCG@3	& nDCG@5 &	ERR@1	& ERR@3& 	ERR@5 \\
\midrule
$Marto_P$ &	52.11	& 50.57& 	37.16& 	35.71& 	25.72	& 22.37	& 22.96	& 13.20& 	16.90& 	20.19 \\
$Marto_B$& 	53.62& 	51.25	& 38.22	& 36.09& 	27.47& 	23.88	& 23.65	& 14.03& 	18.48	& 20.80 \\
$Marto$ & 	\textbf{54.70} & 	\textbf{53.38} & 	\textbf{40.47} & 	\textbf{36.65}& 	\textbf{30.66}& 	\textbf{24.57}	& \textbf{23.81}	& \textbf{15.63} & 	\textbf{20.21}	& \textbf{21.64} \\

   \bottomrule
    \end{tabular}
    \vspace{-1mm}
\end{table*}

\section{Experimental Results of the Intermediate Modules}
\label{intermediate}
For a more comprehensive understanding of \OurModel, we attach the performance score of the image selection module in the following table (see Table \ref{appendixism}). Specifically, we perform human-assessed accuracy evaluation to manually check the porpotion of retrieved images that are accurate and relevant to the question and the topic. Also, we attach the performance of the multimodal-enhanced question classification module in Table \ref{appendixmqc}.  Note that even though our method (T5) does not outperform the Bert model on the question classification subtask on every metric, our overall retrieval system outperforms VisualBert. This is due to the superiority of multi-task learning in the generative framework, where knowledge can better be incorporated across different tasks via different prompts in the generative model. Which also verifies the effectiveness of our generative framework.

\section{Model structure analysis}
\label{modelstructure}
For the format of doc-id, we also compare different identifier generation strategies (See the following Table \ref{appendixdocid}).  Where $Doc_N$ refers to using document number as the identifier, $Doc_{F5}$ refers to using the first 5 words as the identifier. $Doc_K$ is our method, using top-5 keywords as the identifier.

For the reason of choosing VLT5 as the base model, we also performed some experiments. First, VLT5 is among the models that perfectly fit our requirement, taking the image and text as input and outputting the text. We also try other methods and list the results as follows in Table \ref{appendixbackbone}, where $Marto_P$ uses VLP~\cite{Zhou2019UnifiedVP} as backbone, and $Marto_B$ uses VLBart as backbone.

\section{Ethics Statement}
\label{ethics}
When creating the dataset, we take stringent measures to ensure the confidentiality of user intentions (facets) by keeping them strictly hidden from the annotators. Additionally, we prioritize the privacy of the collected images, implementing safeguards to protect sensitive and personal information. To assure that, we will only disclose the image URLs when sharing the dataset.

\section{Ongoing Directions}
\label{ongingdirections}
The effectiveness of \OurModel indicates its potential for future directions, such as exploring the MQC task in a multi-turn scenario and improving image selection methods. 
Furthermore, it is intriguing to investigate how images can enhance the user experience at other interaction stages beyond clarification. 
As image selection is critical, improving methods for selecting images and considering their diversity is vital. Therefore, it's of great research value to explore methods with a more sophisticated structure. Additionally, considering the increasing research interest in large language models (LLMs) for various NLP and IR tasks, we are extending current \OurModel (based on smaller-scaled pretrained model VLT5) to adopt multimodal LLMs as base model, such as BLIP-2~\cite{Li2023BLIP2BL}. By comparing the upgraded \OurModel with existing LLMs such as GPT-4V~\cite{OpenAI2023GPT4TR}, Llama~\cite{Touvron2023LLaMAOA}, we are seeking to see how LLMs will help improve the clarification phase in the search process.

\end{document}